\pdfoutput=1

\documentclass[11pt]{article}

\usepackage[final]{acl}

\usepackage{booktabs}
\usepackage{times}
\usepackage{svg}
\usepackage{float}
\usepackage{latexsym}
\usepackage{amsmath, amssymb} 
\usepackage{mathtools} 
\usepackage{enumitem} 

\usepackage[T1]{fontenc}

\usepackage[utf8]{inputenc}

\usepackage{microtype}

\usepackage{inconsolata}

\usepackage{graphicx}
\usepackage{natbib}
\title{Igniting Creative Writing in Small Language Models: LLM-as-a-Judge versus Multi-Agent Refined Rewards}
\date{\today}

\author{
  Xiaolong Wei$^{1}$$^*$,
  Bo Lu$^{2}$$^*$,
  Xingyu Zhang$^{3}$,
  Zhejun Zhao$^{2}$$^{\dag}$ \\
  {\bf Dongdong Shen}$^{2}$,
  {\bf Long Xia}$^{2}$,
  {\bf Dawei Yin}$^{2}$ \\
  $^{1}$Beihang University \ \ 
  $^{2}$Baidu Inc. \\
  $^{3}$Beijing Jiaotong University \\
  \texttt{xiaolongwei@buaa.edu.cn, zhaozhejun@baidu.com}
}

\begin{document}
\maketitle

\def\thefootnote{*}\footnotetext{Co-first authors with equal contributions.}
\def\thefootnote{\dag}\footnotetext{Corresponding author}

\begin{abstract}
Large Language Models (LLMs) have demonstrated remarkable creative writing capabilities, yet their substantial computational demands hinder widespread use. Enhancing Small Language Models (SLMs) offers a promising alternative, but current methods like Supervised Fine-Tuning (SFT) struggle with novelty, and Reinforcement Learning from Human Feedback (RLHF) is costly. This paper explores two distinct AI-driven reward strategies within a Reinforcement Learning from AI Feedback (RLAIF) framework to ignite the creative writing of a 7B-parameter SLM, specifically for generating Chinese greetings. The first strategy employs a RM trained on high-quality preference data curated by a novel multi-agent rejection sampling framework designed for creative tasks. The second, more novel strategy utilizes a principle-guided LLM-as-a-Judge, whose reward function is optimized via an adversarial training scheme with a reflection mechanism, to directly provide reward signals. Comprehensive experiments reveal that while both approaches significantly enhance creative output over baselines, the principle-guided LLM-as-a-Judge demonstrably yields superior generation quality. Furthermore, it offers notable advantages in training efficiency and reduced dependency on human-annotated data, presenting a more scalable and effective path towards creative SLMs. Our automated evaluation methods also exhibit strong alignment with human judgments. Our code and data are publicly available at \href{https://github.com/weixiaolong94-hub/Igniting-Creative-Writing-in-Small-Language-Models}{Github}.
\end{abstract}

\section{Introduction}
Creative writing, a cornerstone of human expression and communication \citep{kaufmann2012importance, bakar2021qualities}, intrinsically demands not only literary merit and emotional resonance but also a significant degree of personalization to effectively engage its audience \citep{bakar2021qualities}. While users increasingly turn to online platforms for creative inspiration, existing retrieval-based methods often fall short in delivering content that is sufficiently tailored to individual needs and contexts, a limitation that has become more pronounced with the advent of advanced generative models. This underscores a growing demand for generative systems capable of producing context-aware, responsive, and personalized creative text \citep{richardson2023integrating}.

The advent of Large Language Models (LLMs) such as GPT-4o \citep{hurst2024gpt} and DeepSeek-V3 \citep{liu2024deepseek} has revolutionized text generation, demonstrating remarkable capabilities in creative writing tasks. However, under high request volumes, the substantial computational footprint and high inference latency of these large-scale models present significant barriers to their widespread deployment and practical application. Consequently, enhancing Small Language Models (SLMs, typically <10B parameters), such as the Qwen2.5 7B model we employ \citep{team2024qwen2}, to achieve comparable creative prowess while maintaining efficiency has become a critical research frontier. This pursuit aligns with broader trends where modern applications increasingly prioritize dynamic content personalization \citep{li2025multi, cui2025diffusion, cui2025multi} while also emphasizing information's expressiveness and reliability \citep{tong2024mmdfnd, lu2025dammfnd, zeng2025futuresightdrive}. It is crucial to note that generic, un-fine-tuned SLMs often lack the sophisticated generative abilities required for high-quality creative writing \citep{gomez2023confederacy}.

Prevailing methodologies for enhancing SLMs predominantly involve Supervised Fine-Tuning (SFT) and Reinforcement Learning from Human Feedback (RLHF) \citep{ouyang2022training}. While SFT can effectively adapt SLMs to specific styles, it often struggles to foster genuine novelty and generalization \citep{zhou2023lima, sanh2021multitask}—attributes paramount for compelling creative writing. RLHF, on the other hand, relies on high-quality reward models typically trained on extensive human preference data, the annotation of which is labor-intensive and expensive \citep{ziegler2019fine}.

To surmount these limitations, we investigate two distinct reward strategies:
\begin{itemize}
    \item \textbf{A Refined Reward Model:} We develop an RM trained on meticulously curated preference data. This data is generated and filtered by a novel multi-agent framework designed to ensure high quality and relevance for creative tasks.
    \item \textbf{Principle-Guided LLM-as-a-Judge:} Drawing inspiration from "LLM-as-a-Judge" paradigms \citep{zheng2023judging}, we directly employ a powerful LLM as the reward provider. Crucially, this LLM's judgments are guided by explicitly defined creative writing principles and its reward function is further optimized via an adversarial training scheme \citep{wang2024llm}.
\end{itemize}

We conduct comprehensive experiments on generating Chinese greetings using 7B-parameter SLMs, specifically the Qwen2.5 7B model. Our findings reveal that while both RL-based approaches significantly enhance creative output compared to baselines, the principle-guided LLM-as-a-Judge strategy yields demonstrably superior results in terms of generation quality. These outcomes are rigorously validated through both human evaluations and LLM-based assessments, including an analysis of their alignment. Furthermore, the LLM-as-a-Judge approach exhibits notable advantages in training efficiency and reduced dependency on human-annotated data.

Our main contributions are threefold:
\begin{itemize}
    \item We introduce a novel principle-guided LLM-as-a-Judge reward mechanism, optimized adversarially, for effectively steering RL towards enhancing SLM creative writing capabilities.
    \item We propose a multi-agent framework for generating and filtering high-quality preference data, enabling the training of more effective reward models for creative domains.
    \item We present a systematic comparison of these two reward paradigms for SLM-based creative writing, corroborated by extensive LLM-based and human evaluations, and offer insights into their alignment and practical trade-offs.
\end{itemize}

\section{Related Work}
The landscape of artificial intelligence in creative writing has been dramatically reshaped by LLMs. These models, such as the GPT series \citep{brown2020language, achiam2023gpt} and LLaMA \citep{touvron2023llama}, trained on vast text corpora, demonstrate unprecedented capabilities in generating diverse creative texts, including complex narratives, poetry, and scripts, exhibiting high fluency, style adaptation, and thematic coherence. Researchers have developed techniques like planning \citep{yang2022re3}, controllable generation \citep{li2022diffusion}, and structured decomposition frameworks like Branch-Solve-Merge \citep{saha2023branch} to further enhance and guide LLMs' creative output.

Beyond autonomous generation, recent work increasingly focuses on LLMs as co-creative partners for human writers, exploring interaction dynamics for tasks such as brainstorming and outlining \citep{gero2023social}. The concept of multi-agent systems collaborating on writing tasks is also an emerging area.

Despite these advancements in generation capabilities, evaluating the creativity of LLM-produced text remains a complex challenge \citep{chakrabarty2024art, kim2025evaluating}. Traditional automatic metrics are insufficient for capturing subjective qualities like originality and emotional depth. To address this, recent work has explored Self-Rewarding Language Models \citep{yuan2024self} that iteratively improve by generating their own training rewards, though automated assessments still do not yet reliably align with human judgments \citep{chakrabarty2024art, li2025automated}.

These persistent challenges in aligning automated evaluation with human judgment highlight fundamental open problems: how to build effective reward signals for training generative models and achieve reliable automated evaluation in this subjective domain.

\newcommand{\Prompts}{\mathcal{P}}
\newcommand{\Responses}{\mathcal{R}}
\newcommand{\DatasetHQ}{\mathcal{D}_{\text{HQ}}} 
\newcommand{\DatasetPref}{\mathcal{D}_{\text{pref}}} 
\newcommand{\RetrievedSet}{E} 
\newcommand{\EvalPos}{\mathcal{E}^{+}} 
\newcommand{\EvalNeg}{\mathcal{E}^{-}} 
\newcommand{\JudgementInitial}{\mathcal{S}_{\text{initial}}} 
\newcommand{\JudgementFinal}{\mathcal{S}_{\text{final}}} 
\newcommand{\DetectorOutputSpace}{\mathcal{S}_{\text{D}}} 
\newcommand{\ReflectorAdvice}{\mathcal{R}_{\text{D}}} 
\newcommand{\TrueLabelSpace}{\mathcal{Y}_{\text{true}}} 

\section{Methodology}
\begin{figure*}[h] 
  \includegraphics[width=\textwidth]{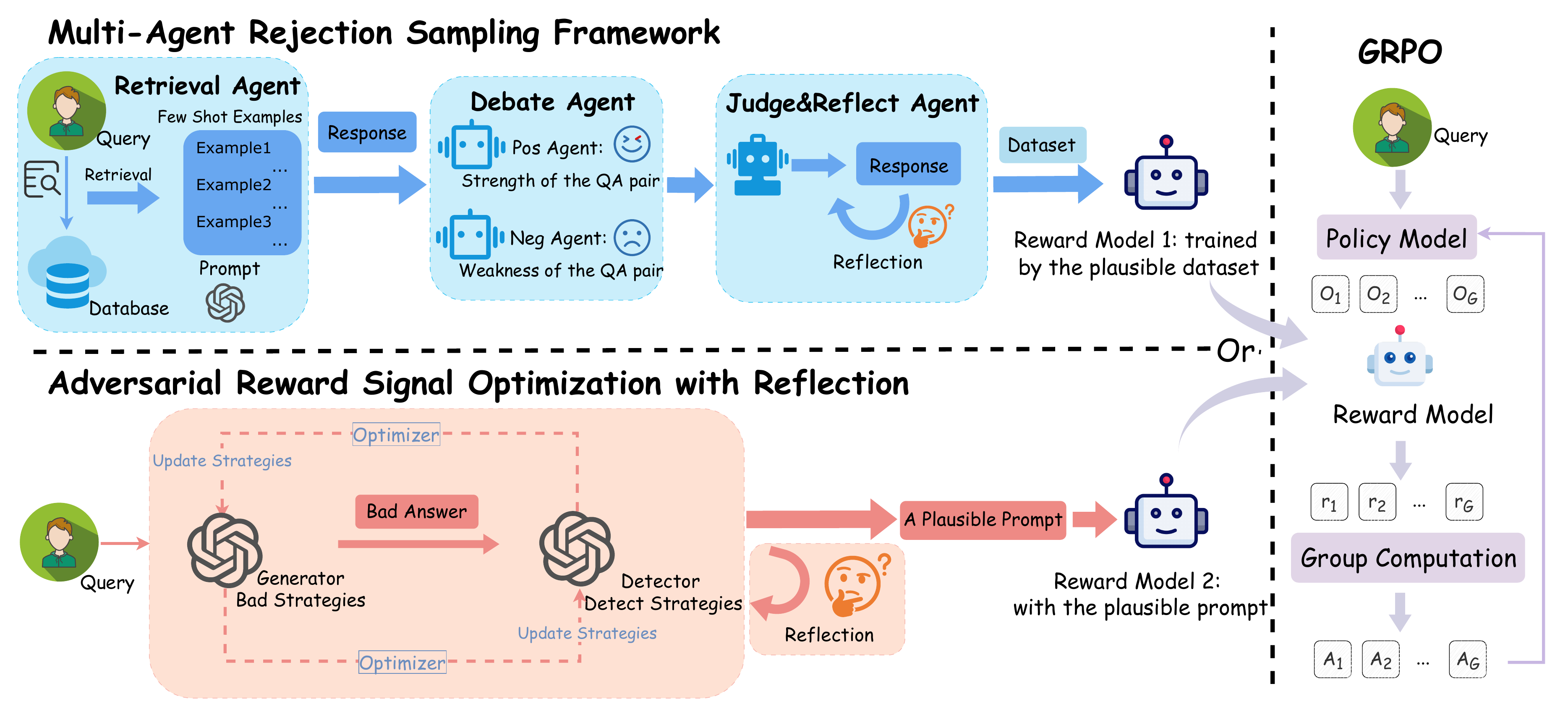} 
  \caption{The figure depicts two distinct reward signals. Signal 1 is derived from a multi-agent system, yielding a reward model. Signal 2 is generated via adversarial interaction (Generator-Detector) and reflection, producing a prompt. Both signals are separately used to train GRPO.}
  \label{fig:flow_chart}
\end{figure*}

To enhance the creative writing capabilities of our target SLM, we employ a RLAIF paradigm. The central tenet of RLAIF is to refine the SLM's policy using reward signals derived from AI-driven evaluations of its generated outputs. Our primary contribution lies in the exploration and comparison of two distinct and sophisticated strategies for generating these crucial reward signals, which are designed to capture the multifaceted nature of creative text. These strategies are: 1) a meticulously refined RM trained on preference data curated by a multi-agent system, and 2) a dynamic reward signal obtained from an adversarially trained, principle-guided LLM acting as a judge (LLM-as-a-Judge). The complete process is detailed in Fig.~\ref{fig:flow_chart}.
In the subsequent sections, we first detail the multi-agent framework for preference data generation and RM training (Section~\ref{sec:multi_agent_eval}). We then describe the adversarial approach for optimizing an LLM-as-a-Judge as a direct reward provider (Section~\ref{sec:adversarial_reward}). Finally, Section~\ref{sec:rlaif} outlines how the reward signals derived from these two strategies are integrated into the RLAIF process to optimize the SLM.

\subsection{Multi-Agent Rejection Sampling Framework}
\label{sec:multi_agent_eval}
The evaluation of LLMs by a single LLM instance, while scalable, can suffer from inherent biases, limited perspectives, and potential instability \cite{zheng2023judging}. To mitigate these challenges, we introduce a multi-agent collaborative evaluation system. This system operationalizes a collaborative paradigm, drawing inspiration from approaches where multiple agents engage in debate or structured discussion to refine assessments and achieve more robust outcomes \cite{chan2023chateval, du2023improving}. By simulating a nuanced, rigorous, and bias-resistant assessment process, our framework aims to leverage the collective intelligence and error-correction capabilities inherent in multi-agent interactions \cite{liang2023encouraging}. This approach aligns with a broader trend in AI systems where complex tasks are decomposed and managed by specialized, collaborative agents to achieve a goal \citep{li2025towards}. The primary output of this system is high-fidelity preference data, denoted as $\mathcal{D}_{\text{pref}}$. This dataset is specifically curated to be suitable for training robust reward models, which can subsequently be employed to filter and rank generated content based on nuanced quality dimensions. All prompts and cases are provided in Appendix~\ref{sec:prompts}.

\subsubsection{Retrieval Agent}
The Retrieval Agent, implementing the function $Rtr: \Prompts \to \mathcal{P}(\DatasetHQ)$, retrieves relevant context for evaluation. Upon receiving an input prompt $p \in \Prompts$, it queries a pre-computed vector index (built from $\DatasetHQ$) using similarity metrics (e.g., cosine similarity on embeddings) to fetch the set $\RetrievedSet = Rtr(p) = \{(p'_j, r'_j)^*\}_{j=1}^{k}$ of $k$ high-quality prompt-response pairs. These pairs serve as few-shot examples, providing contextual grounding and quality benchmarks for the subsequent evaluation agents.

\subsubsection{Debate Agents: Positive and Negative Perspectives}
This module employs two adversarial agents, embodying the functions $f_{\text{pos}}: (\Prompts, \Responses, \RetrievedSet) \to \EvalPos$ and $f_{\text{neg}}: (\Prompts, \Responses, \RetrievedSet) \to \EvalNeg$, to conduct a structured debate on the quality of a given response $r$ for prompt $p$.
\begin{itemize}
    \item \textbf{Positive Agent ($f_{\text{pos}}$):} Identifies and articulates the strengths and merits of the response $r$, such as novelty, coherence, emotional resonance, or alignment with the prompt's intent. Its output is a structured positive evaluation $\varepsilon^+ \in \EvalPos$.
    \item \textbf{Negative Agent ($f_{\text{neg}}$):} Identifies and articulates the weaknesses and potential issues within $r$, such as factual inaccuracies, logical fallacies, stylistic clichés, or lack of creativity. Its output is a structured negative evaluation $\varepsilon^- \in \EvalNeg$.
\end{itemize}
This structured debate mechanism compels a multi-faceted analysis, surfacing both positive and negative aspects that might be overlooked by a single evaluator due to confirmation bias or inherent model preferences.This process yields a more comprehensive and less biased assessment, crucial for subjective domains like creative writing.

\subsubsection{Judge Agent}
The Judge Agent, implementing $f_{\text{judge}}: (\Prompts, \Responses,\EvalPos, \EvalNeg) \to \JudgementInitial$, synthesizes the evaluations $\varepsilon^+$ and $\varepsilon^-$ from the debate agents. It weighs the conflicting arguments, assesses the relative importance of identified strengths and weaknesses, and formulates a holistic initial judgment $S_{\text{initial}} \in \JudgementInitial$. This simulates a reasoned decision-making process based on multifaceted evidence.

\subsubsection{Reflect Agent}
Following the initial judgment, the Reflect Agent, implementing $f_{\text{reflect}}: (\Prompts, \Responses,\JudgementInitial, \EvalPos, \EvalNeg) \to \JudgementFinal$, performs a critical review of $S_{\text{initial}}$ and the supporting arguments $\varepsilon^+$ and $\varepsilon^-$. It scrutinizes the Judge Agent's reasoning for logical consistency and completeness. If flaws are detected, the Reflect Agent may override $S_{\text{initial}}$ and potentially trigger a re-evaluation. Otherwise, it ratifies the initial judgment, resulting in the final assessment $S_{\text{final}} \in \JudgementFinal$. This reflection step enhances the reliability and robustness of the final evaluation. Based on $S_{\text{final}}$, a preference pair $(p, r_{\text{chosen}}, r_{\text{rejected}})$ is determined and added to the preference dataset $\DatasetPref$.

\subsection{Adversarial Reward Signal Optimization with Reflection}
\label{sec:adversarial_reward}
Inspired by Generative Adversarial Networks (GANs) and related approaches like LLM-GAN \cite{wang2024llm}, we propose an adversarial framework to dynamically generate and refine reward signals for RL-based policy optimization. This framework comprises a Generator, a Detector, and a novel Reflector component. Further details are provided in Appendix~\ref{sec:detail_of_gan}.

\subsubsection{Generator-Detector Adversarial Dynamics}
\begin{itemize}
    \item \textbf{Generator ($\pi_G$):} The Generator, parameterized by $\theta_G$, aims to produce responses $r$ for a given prompt $p$ according to its policy $\pi_G(r|p; \theta_G)$. Its goal is to generate bad responses that are hard to distinguish.
    \item \textbf{Detector ($f_D$):} The Detector, parameterized by $\theta_D$, acts as a discriminator. It learns to distinguish responses $r$ generated by $\pi_G$. It assigns a score $f_D(p, r; \theta_D) \in \{0, 1\}$, where 1 represents a good response and 0 represents a bad response.
\end{itemize}
These components engage in adversarial training. The Detector is trained to maximize its ability to correctly classify responses, while the Generator aims to produce indistinguishable bad responses to
deceive the Detector.

\subsubsection{Reflector-Enhanced Detector Optimization}
To further improve the Detector's reliability, we introduce the Reflector module ($f_{\text{Rf}}: (\Prompts, \Responses, \DetectorOutputSpace, \TrueLabelSpace) \to \ReflectorAdvice$). When the Detector $f_D$ misclassifies a response $(p, r)$ compared to a reference label $y_{\text{true}} \in \TrueLabelSpace$ (where $y_{\text{true}}$ could indicate if $r$ is genuinely high-quality or not, obtained from $\DatasetPref$ or human annotation), the Reflector is activated.
The Reflector analyzes the triplet $(p, r, s_D=f_D(p, r; \theta_D))$ alongside $y_{\text{true}}$ to diagnose the cause of the Detector's error. Based on this analysis, it generates structured feedback or advice $R_D \in \ReflectorAdvice$. This advice $R_D$ can be used to guide the Detector's optimization process (e.g., "Increase weight on detecting emotional flatness"). This explicit reflection mechanism allows the Detector to learn from its mistakes beyond the implicit adversarial signal, improving its robustness and alignment with desired quality criteria.

\subsection{RLAIF for Creative Writing Enhancement}
\label{sec:rlaif}
This section details the integration of the previously described AI-generated reward signals into the RLAIF process. Our goal is to optimize the target SLM, Qwen2.5-7B-Instruct, for enhanced creative writing proficiency by leveraging nuanced feedback. We investigate two primary sources for the reward signal used within the RLAIF process:
\begin{itemize}
    \item \textbf{Multi-Agent Preference Reward Model (RM):} A reward model $R_{\text{MA}}(p, r; \phi_{\text{RM}})$ is trained on the high-quality preference dataset $\DatasetPref$ generated by the multi-agent evaluation system described in Section~\ref{sec:multi_agent_eval}. The RM learns to predict the preferences expressed in $\DatasetPref$, typically using a loss function like:
\begin{equation}
\label{eq:rm_loss_option2a}
\begin{multlined}[b][0.9\columnwidth] 
\mathcal{L}_{\text{RM}} = -\mathbb{E}_{(p, r_c, r_r) \sim \mathcal{D}_{\text{pref}}} \\
\shoveleft{\left[ \log \sigma \left( R_{\text{MA}}(p, r_c; \phi_{\text{RM}}) - R_{\text{MA}}(p, r_r; \phi_{\text{RM}}) \right) \right]}
\end{multlined}
\end{equation}

    where $\sigma$ is the sigmoid function. The output $R_{\text{MA}}(p, r)$ serves as the reward signal.
    \item \textbf{Adversarial Detector Reward Signal:} The output score $s_D = f_D(p, r; \theta_D)$ from the adversarially trained and reflector-enhanced Detector (detailed in Section~\ref{sec:adversarial_reward}) is used directly as a reward signal, $R_D(p, r) = f_D(p, r; \theta_D)$. This signal reflects the response's ability to meet the criteria implicitly learned by the dynamic LLM-based judge.
\end{itemize}

 We apply GRPO algorithm\citep{shao2024deepseekmath} to optimize the Qwen2.5-7B-Instruct. The advantage $A_t$ is calculated based on trajectories sampled from the policy $\pi_\theta$ and rewards obtained from either $R_{\text{MA}}$ or $R_D$. We compare the effectiveness of these distinct reward mechanisms in enhancing the models' creative writing capabilities across various dimensions.

\begin{table*}[h!] 
    \centering
    \begin{tabular}{l c c c c}
        \toprule
        & Accuracy & Precision & Recall & F1-score \\
        \midrule
        Multi-Agent Framework & 87.60\% & 87.38\% & 87.90\% & 0.8764 \\
        Adversarial Framework & 85.50\% & 78.54\% & 97.70\% & 0.8708 \\

        \bottomrule
    \end{tabular}
    \caption{Comparison of two different frameworks on the evaluation set.}
    \label{tab:Comparison_two_frameworks} 
\end{table*}

\begin{table*}[h!] 
    \centering
    \begin{tabular}{l c c c}
        \toprule
        & Signal-1 & Signal-2 & Human \\
        \midrule
        GPT-4o & 49.0\% & 46.8\% & 50.0\% \\
        Ernie-4.5 & 76.4\% & 88.2\% & 87.6\% \\
        DeepSeek-V3 & 91.0\% & 94.2\% & 93.0\% \\
        \hline
        Qwen2.5-7B-Instruct & 59.2\% & 56.0\% & 57.6\% \\
        SFT + Qwen2.5-7B-Instruct & 92.0\% & 92.6\% & 90.0\% \\
        Reward Model + RL & - & - & - \\
        LLM-as-a-Judge + RL & \textbf{92.4\%} & \textbf{96.6\%} & \textbf{95.0\%} \\
        SFT + Reward Model + RL & 92.2\% & 96.0\% & 94.6\% \\
        SFT + LLM-as-a-Judge + RL & 89.6\% & 96.0\% & 93.0\% \\

        \bottomrule
    \end{tabular}
    \caption{Comparison of the excellence rate of the Model under different evaluation mechanisms. This data represents the inference results of the model under high-frequency greetings (for example, Chinese New Year greetings). Here, Signal-1 refers to Section~\ref{sec:multi_agent_eval}, Signal-2 refers to Section~\ref{sec:adversarial_reward}, and Human refers to the evaluation by human experts. Furthermore, the Reward Model + RL method is excluded from the evaluation due to its training not converging.}
    \label{tab:not_ood} 
\end{table*}

\begin{table*}[h!] 
    \centering
    \begin{tabular}{l c c c}
        \toprule
        & Signal-1 & Signal-2 & Human \\
        \midrule
        GPT-4o & 47.6\% & 45.6\% & 50.4\% \\
        Ernie-4.5 & 72.0\% & 81.2\% & 83.0\% \\
        DeepSeek-V3 & 74.0\% & 83.8\% & 85.6\% \\
        \hline
        Qwen2.5-7B-Instruct & 47.6\% & 52.8\% & 53.8\% \\
        SFT + Qwen2.5-7B-Instruct & 80.2\% & 85.2\% & 86.2\% \\
        Reward Model + RL & - & - & - \\
        LLM-as-a-Judge + RL & \textbf{91.0\%} & \textbf{93.4\%} & \textbf{92.4\%} \\
        SFT + Reward Model + RL & 89.4\% & 90.6\% & 91.2\% \\
        SFT + LLM-as-a-Judge + RL & 85.0\% & 89.0\% & 90.2\% \\

        \bottomrule
    \end{tabular}
    \caption{Comparison of the excellence rate of the Model under different evaluation mechanisms. This data represents the inference results of the model under ordinary greetings(for example, greetings for a new car). Furthermore, the Reward Model + RL method is excluded from the evaluation due to its training not converging.}
    \label{tab:ood} 
\end{table*}
\section{Experiments}
\subsection{Task Design}
This study centers on enhancing the generation of Chinese greetings. These greetings are prevalent in Chinese culture for significant festivals like the Spring Festival and Mid-Autumn Festival, indicating a high practical demand and rich contextual nuances. This specific focus allows for an in-depth exploration of creative text generation within a culturally significant and frequently utilized domain. The details are provided in Appendix~\ref{sec:scope_and_characteristics}.

\subsection{Datasets}
\label{sec:datasets}
Our experiments leverage several datasets constructed for distinct purposes: training a retrieval-augmented multi-agent system, developing reward models, fine-tuning the policy model via RLAIF, and comprehensive final evaluation. All data was sourced from online interactions related to Chinese greetings, with meticulous preprocessing to remove Personally Identifiable Information (PII). Specific business-related source details remain desensitized.

\paragraph{Retrieval Corpus}
To equip our multi-agent evaluation system (Section~\ref{sec:multi_agent_eval}) with high-quality contextual examples, we curated a retrieval corpus comprising 23,442 instances. These instances were selected from a larger online collection based on their high user click-through rates and frequent replication, indicative of their perceived quality and relevance.

\paragraph{Reward Model Training Data}
For training the preference-based reward model, we initially collected 10,000 user queries from online sources. These queries, along with candidate responses, were processed through our multi-agent rejection sampling framework. This procedure yielded 7,896 preference pairs, each structured as $(query, response_{chosen}, response_{rejected})$. This dataset was then partitioned into an 80\% training set and a 20\% held-out test set for RM development.

\paragraph{Policy Optimization (GRPO) Data}
A separate set of 4,000 distinct online queries was utilized for fine-tuning the target SLM using the GRPO algorithm. This dataset was also divided into an 80:20 train/test split to guide the RLAIF process.

\paragraph{Final Evaluation Set}
To rigorously assess the performance of all compared models, we constructed a dedicated evaluation set of 2,000 query-response pairs. This set was carefully balanced, containing 1,000 "high-quality" instances (heuristically labeled '1'), selected from data exhibiting high click-through and replication rates, and 1,000 "low-quality" instances (labeled '0'), derived from data with lower engagement metrics. This dataset serves as the primary benchmark for both our automated and human evaluations.
\subsection{Rubric Design}
\label{sec:evaluation_metrics}
The evaluation rubric provides a holistic view of greetings quality, comprising five dimensions with respective weights: Language Quality (30\%), Creativity (30\%), Emotional Resonance (15\%), Cultural Appropriateness (15\%), and Content Richness (10\%).

\textbf{Language Quality} assesses fluency and precision. Essential for effective communication, its importance in NLG systems is well-recognized \cite{van2019best, que2024hellobench}, with modern approaches using LLMs for nuanced assessment \cite{liu2023g} and considering aspects like style and meaning preservation \cite{chim2025evaluating}.

\textbf{Creativity} evaluates the generation of innovative elements like unique metaphors or novel perspectives, distinguishing memorable greetings. This involves producing novel, surprising, and valuable outputs \cite{zhang2025noveltybench}, crucial for pushing NLG beyond mere replication \cite{eldan2023tinystories, ismayilzada2024evaluating, peng2022controllable}.

\textbf{Emotional Resonance} measures the capacity to evoke strong feelings or genuine connection. This is vital as greetings are inherently emotional, and the text's ability to connect on an emotional level is key \cite{cao2025does, li2022emotion, ruhlemann2024effect}.

\textbf{Cultural Appropriateness} ensures alignment with the specific cultural context, respecting social norms, traditions, and event-specific sensitivities \cite{li2024culture}. There's growing emphasis on developing culturally sensitive models that avoid biases \cite{pawar2024survey, naous2025origin, naous2024having}.

\textbf{Content Richness} ensures greetings convey sufficient emotional depth and personalized information concisely. It emphasizes meaningful, relevant, and comprehensive content within a brief format, delivering value and substance \cite{gao2025llm, zheng2023judging, nimah2023nlg}.

Each dimension is rated on a discrete scale from 1 to 3 points. A final aggregate score is computed as a weighted average. Based on this, a binary classification is performed: acceptable (label 1) if the total weighted score is $\ge 2$, and unacceptable (label 0) otherwise.

\subsection{Implementation Details}
The reward model in this study is implemented using the Llama Factory framework \citep{zheng2024llamafactory} and fine-tuned with the LoRA method \citep{hu2022lora}. We train a scalar reward model~$R_\theta$ by adding a single linear value head to the backbone LLM and fine-tuning it on human preference
pairs $(x,y^{+},y^{-})$ with the Bradley–Terry loss
$\mathcal{L}=-\log\sigma\!\bigl(R_\theta(x,y^{+})-R_\theta(x,y^{-})\bigr)$,
following \citet{stiennon2020learning} and \citet{ouyang2022training}. Further details are provided in Appendix~\ref{sec:implement_details}.

\begin{table*}[t] 
    \centering
    \begin{tabular}{l c c c c}
        \toprule
        & Accuracy & Precision & Recall & F1-score \\
        \midrule
        \textbf{Multi-Agent Framework (Full)} & 87.60\% & 87.38\% & 87.90\% & 0.8764 \\
        w/o Positive Agent & 53.65\% & 98.67\% & 7.40\% & 0.1377 \\
        w/o Negative Agent & 50.05\% & 50.03\% & 100.00\% & 0.6669 \\
        w/o Judge Agent & 81.45\% & 76.19\% & 91.50\% & 0.8314 \\
        w/o Reflect Agent & 76.40\% & 75.83\% & 77.50\% & 0.7666 \\
        \textbf{Adversarial Framework (Full)} & 85.50\% & 78.54\% & 97.70\% & 0.8708 \\
        w/o Reflect Agent & 81.00\% & 68.49\% & 99.10\% & 0.8100 \\
        \bottomrule
    \end{tabular}
    \caption{Ablation study of different agents.}
    \label{tab:ablation_study} 
\end{table*}

\section{Results and Discussion}

\subsection{RQ1: Can LLMs achieve alignment with human evaluation?}
Following the evaluation criteria detailed in Section~\ref{sec:evaluation_metrics}, we engaged a team of professionally trained evaluators to assess the generated greetings across five dimensions: language quality, creativity, emotional resonance, cultural appropriateness, and content richness. All evaluators were of Chinese nationality and ethnicity, residing and working in China. The team comprised graduate-level educated interns and full-time employees, all of whom were compensated for their work. Each dimension was scored independently by multiple annotators from this team to ensure reliability.

\begin{figure}[h]
  \includegraphics[width=\columnwidth]{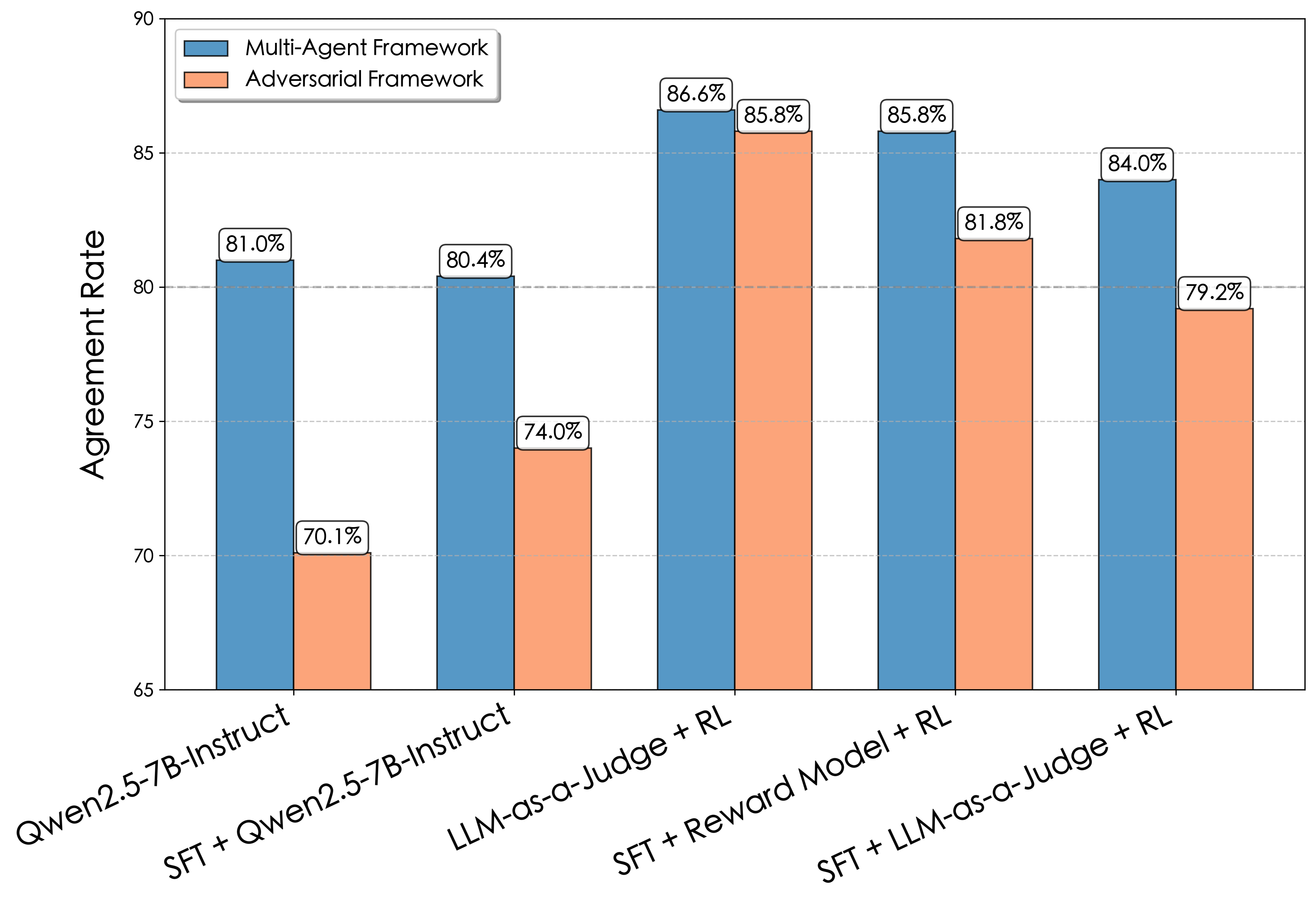}
  \caption{Comparison of agreement rate between different models and human under two evaluation frameworks.}
  \label{fig:agreement_rate}
\end{figure}

Fig.~\ref{fig:agreement_rate} illustrates the agreement rates between human evaluations and two proposed automatic evaluation frameworks: Multi-Agent Framework and Adversarial Framework. As depicted, both the Multi-Agent Framework and Adversarial Framework approaches demonstrate substantial agreement with human judgments, consistently exceeding 70\% across different models. This strong correlation provides compelling evidence for the effectiveness of our proposed mechanisms in approximating human evaluation, thereby offering a potential solution to the time-consuming nature and high cost associated with extensive human annotation.

Furthermore, a closer examination of Fig.~\ref{fig:agreement_rate} reveals that the Multi-Agent Framework exhibits a higher degree of alignment with human evaluators, achieving agreement rates ranging from 80\% to 87\% across the evaluated models. This excellent performance suggests that through the Multi-Agent Framework, it is possible to more accurately identify the strengths and weaknesses of greetings and more closely align with human evaluations of greetings.

In summary, both the Multi-Agent Framework and Adversarial Framework proposed in this work demonstrate a significant capacity for aligning with human assessments. This alignment offers a promising avenue for substantially alleviating the burden of manual evaluation in the context of generative text tasks.

\subsection{RQ2: Efficacy of Reward Model + RL in Enhancing Creative Writing}

Tab.~\ref{tab:not_ood} and Tab.~\ref{tab:ood} present a comparative evaluation of mainstream LLMs against our models trained using distinct methodologies. The primary evaluation metric is the excellence rate (1 indicating positive, 0 negative assessment) across predefined dimensions. Specifically, Tab.~\ref{tab:not_ood} showcases performance on greetings from high-frequency user queries, while Tab.~\ref{tab:ood} evaluates those from ordinary queries.

The results consistently demonstrate that a pipeline employing SFT followed by Reward Model training and RL significantly outperforms the SFT-only baseline across both high-frequency and ordinary query scenarios. For instance, as detailed in Tab.~\ref{tab:ood}, the SFT+RM+RL approach yields substantial improvements, achieving gains of 11.5\% on the Signal-1 dimension, 6.3\% on Signal-2, and 5.8\% on the human evaluation dimension.

Furthermore, the SFT+RM+RL trained models surpass several contemporary mainstream LLMs on both query types. These findings provide compelling evidence for the efficacy of integrating RM and RL techniques in enhancing creative writing capabilities, particularly for generating contextually relevant and high-quality greetings. This improvement indirectly validates our multi-agent based data filtering strategy for RM training, which contributes to the superior performance observed in the downstream generation task.

\subsection{RQ3: Does "LLM-as-a-Judge" offer advantages over other reward signals?}
A fundamental distinction differentiates the reward signals from LLM-as-a-Judge and conventional RMs. LLM-as-a-Judge provides a binary (0 or 1) reward, a discrete signal, while RMs generate continuous values, offering fine-grained feedback.

Empirical results (Tab.~\ref{tab:not_ood}) demonstrate that the LLM-as-a-Judge + RL approach achieves state-of-the-art (SOTA) performance, with excellence rates of 92.4\%, 96.6\%, and 95.0\% across three distinct evaluation metrics. This performance surpasses several contemporary mainstream LLMs (e.g., GPT-4o, Ernie-4.5, DeepSeek-V3). While Tab.~\ref{tab:ood} indicates a marginal decrease on ordinary queries, the LLM-as-a-Judge + RL method maintains SOTA results. Further details regarding the LLM-as-a-Judge + RL training process are provided in Section~\ref{sec:analysis_of_training}.

These findings compellingly affirm the efficacy of LLM-as-a-Judge + RL in augmenting creative writing capabilities for both high-frequency and ordinary queries, generally outperforming the RM + RL paradigm. This underscores the potential of discrete reward signals to drive substantial performance gains in RL.

Conversely, training an RM using multi-agent filtered data is notably more complex and resource-intensive. This process requires sequential operation of Retrieval, Positive, Negative, and Reflect Agents for data curation, posing significant temporal and computational overhead, which can impede real-world deployment.

LLM-as-a-Judge presents a more direct and efficient alternative. It leverages Adversarial Reward Signal Optimization, wherein a generator and detector engage in adversarial training to iteratively refine an optimal evaluation prompt. This optimized prompt is then directly used to assess generated content quality. Compared to the intricate RM training pipeline, LLM-as-a-Judge markedly reduces procedural complexity. Consequently, LLM-as-a-Judge offers a more streamlined and advantageous approach for deriving effective reward signals for reinforcement learning in this context.

\section{Ablation Study}
To validate the effectiveness of the key components within our proposed architectures, we conducted a comprehensive ablation study on both the Multi-Agent and Adversarial frameworks. The results, presented in Table~\ref{tab:ablation_study}, systematically quantify the contribution of each module by evaluating the performance of the framework after its removal.

For the \textbf{Multi-Agent Framework}, the ablation study underscores the indispensable role of each agent. The most significant performance degradation is observed upon the removal of the debate agents. Without the Positive Agent, the framework becomes excessively critical, achieving high precision but causing a catastrophic drop in recall to 7.40\%, as it fails to recognize valid positive instances. Conversely, removing the Negative Agent renders the system overly lenient, with recall reaching 100\% at the cost of a near-random precision of 50.03\%. This demonstrates that the adversarial debate mechanism is the cornerstone of the framework, ensuring a multi-faceted and balanced assessment. Furthermore, the removal of the Judge Agent and Reflect Agent also leads to notable performance drops. Notably, the absence of the Reflect Agent results in a more substantial decline in both accuracy and F1-score, suggesting that the final self-correction and ratification step is paramount for ensuring the reliability of the preference data.

In the \textbf{Adversarial Framework}, we investigated the contribution of the reflection mechanism. As shown in Table~\ref{tab:ablation_study}, removing the Reflect Agent causes a significant drop across all metrics, with the F1-score falling from 0.8708 to 0.8100. The Reflect Agent provides crucial supervised feedback when the Detector misclassifies a response, allowing it to learn from its mistakes beyond the implicit adversarial signal from the Generator. This component is vital for grounding the Detector's learning process with ground-truth examples, enhancing its overall robustness and accelerating its alignment with the desired quality criteria.

\section{Conclusion}
In this work, we addressed the challenge of enhancing the creative writing capabilities of SLMs by investigating two distinct AI-generated reward paradigms for RLAIF: a refined RM trained on data from a multi-agent system, and a principle-guided, adversarially-optimized LLM-as-a-Judge.
Our contributions are threefold:
First, we introduced a novel principle-guided LLM-as-a-Judge reward mechanism, optimized adversarially with reflection, which effectively steers RL towards enhancing SLM creative writing.
Second, we proposed a multi-agent framework for generating and filtering high-quality preference data, enabling the training of more effective reward models for creative domains.
Third, through systematic comparison on the task of generating Chinese greetings with 7B SLMs, we demonstrated that both AI-feedback approaches significantly improve creative output. Crucially, the LLM-as-a-Judge strategy not only achieved state-of-the-art generation quality, surpassing both the refined RM approach and strong LLM baselines, but also exhibited greater training efficiency and reduced reliance on expensive human annotations.
Our findings underscore the potential of AI-driven feedback, particularly the dynamic and principle-guided LLM-as-a-Judge, to unlock creative capabilities in more compact and efficient language models, paving the way for broader practical applications. The strong alignment observed between our automated evaluation metrics and human judgments further supports the viability of these approaches.

\section{Limitations}
While our findings are promising, this study has several limitations:
\begin{itemize}
    \item \textbf{Task and Language Specificity:} Our experiments focused on generating Chinese greetings. The generalizability of our findings to other creative writing tasks (e.g., long-form storytelling, poetry, scriptwriting) and other languages, particularly those with different linguistic structures or cultural nuances, requires further investigation.
    \item \textbf{Scale of SLMs:} We concentrated on 7B-parameter SLMs. The effectiveness and scalability of the proposed reward mechanisms for significantly smaller or moderately larger SLMs remain to be explored.
    \item \textbf{Subjectivity of Creativity and Principles:} "Creativity" is inherently subjective. While our rubric and multi-faceted evaluation attempt to capture key aspects, the "principles" guiding the LLM-as-a-Judge, though explicitly defined, might still embed certain biases or perspectives on creativity. The optimal set of principles for diverse creative tasks is an open research question.
    \item \textbf{Complexity of Multi-Agent System:} Although the LLM-as-a-Judge approach is more efficient overall, the multi-agent framework for curating preference data for the refined RM, while effective, introduces its own layer of complexity in terms of design and operation.
    \item \textbf{Depth of Reflection:} The reflection mechanism in the LLM-as-a-Judge's adversarial training and in the multi-agent framework is currently based on LLM analysis. The depth and impact of this reflection, and how to systematically improve its error-correction capabilities, are areas for future work.
    \item \textbf{Potential Risk: Reinforcement of Biases:} The principles guiding the LLM-as-a-Judge or the preference data curated by the multi-agent system may unknowingly encapsulate societal or cultural biases. The RLAIF process could then amplify these biases in the SLM's creative outputs, leading to stereotypical or unfair representations.
\end{itemize}
Future research could address these limitations by exploring broader task domains, diverse languages, different model scales, and more sophisticated methods for defining and adapting creative principles for the LLM-as-a-Judge.

\bibliography{main}

\appendix

\section{Appendix}
\label{sec:appendix}

\subsection{Hyperparameters}
\label{sec:implement_details}
We configure the training with a \texttt{per\_device\_train\_batch\_size} of $16$, \texttt{gradient\_accumulation\_steps} of $8$, a \texttt{num\_train\_epochs} of $5.0$, a \texttt{lora\_rank} of $16$, and a \texttt{warmup\_ratio} of $0.1$. The learning rate (\texttt{learning\_rate}) is set to $2.0 \times 10^{-4}$. The finetuning type (\texttt{finetuning\_type}) is \texttt{lora}, the LoRA target (\texttt{lora\_target}) is \texttt{all}, the learning rate scheduler type (\texttt{lr\_scheduler\_type}) is \texttt{cosine}, \texttt{bf16} is set to \texttt{true}, and the \texttt{ddp\_timeout} is $180000000$. Our experiments are conducted on a system equipped with four NVIDIA A100 GPUs, each with 80GB of memory.

Training of the GRPO model is conducted using the Verl framework. We configure the training with a \texttt{train\_batch\_size} of $32$, a \texttt{max\_prompt\_length} of $256$, and a \texttt{max\_response\_length} of $512$. The learning rate (\texttt{lr}) is set to $3 \times 10^{-7}$. For the KL divergence loss, \texttt{use\_kl\_loss} was \texttt{True}, the coefficient (\texttt{kl\_loss\_coef}) is $0.001$, and the type (\texttt{kl\_loss\_type}) is \texttt{low\_var\_kl}. The entropy coefficient (\texttt{entropy\_coeff}) is $0$. The model is trained for $5$ epochs. Our experiments are conducted on a system equipped with four NVIDIA A100 GPUs, each with 80GB of memory.

\subsection{Scope and Characteristics of Chinese Greetings}
\label{sec:scope_and_characteristics}
This study focuses on enhancing the generation of Chinese greetings. These greetings are deeply embedded in Chinese culture, serving as more than mere pleasantries; they are expressions of good will, aspirations, and the reinforcement of social bonds during times of significant cultural importance. The scope of these greetings is broad, encompassing well wishes for individuals, families, and even businesses, reflecting the holistic nature of festive celebrations.

The characteristics of these greetings are multifaceted:

\begin{itemize}
    \item \textbf{Thematic Focus:} Greetings are heavily themed around the core values and significance of each festival. For Spring Festival, common themes include prosperity and wealth, happiness and well-being, health, and success in endeavors. Mid-Autumn Festival greetings, on the other hand, emphasize family reunion and harmony, well-being, and a fruitful harvest.
    \item \textbf{Auspicious Language:} The language used is highly auspicious and positive, employing phrases and characters associated with good fortune, abundance, and success. This often involves the use of four-character idioms and other set phrases that carry rich cultural meanings.
    \item \textbf{Contextual Variation:} While core themes exist, the specific wording and focus of greetings can vary depending on the recipient (e.g., elders, peers, colleagues), the relationship between the sender and recipient, and the specific regional customs. Greetings exchanged within families might be more personal and intimate than those sent to business associates.
    \item \textbf{Cultural Symbolism:} Greetings frequently incorporate cultural symbols associated with the festival. For Spring Festival, this includes references to the zodiac animal of the year, red envelopes, and items symbolizing luck and prosperity. For Mid-Autumn Festival, the moon and mooncakes, symbolizing reunion and completeness, are central to the greetings.
    \item \textbf{Formulaic yet Flexible:} Many greetings utilize established formulaic expressions, making them instantly recognizable and culturally appropriate. However, there is also a degree of flexibility that allows for personalization and creative variation, particularly in informal contexts or in contemporary digital communication.
    \item \textbf{Performative Aspect:} The act of giving and receiving greetings is a significant social ritual that reinforces relationships and community ties. Whether delivered in person, through cards, or via digital messages, the performance of the greeting is as important as the linguistic content.
\end{itemize}

These characteristics highlight the complexity and cultural depth embedded within Chinese greetings, making their accurate and creative generation a challenging yet rewarding task with significant practical applications.

\begin{table*}[h] 
    \centering
    \begin{tabular}{l c c c c c}
        \toprule
        & Language & Creativity & Emotion & Cultural & Content\\
        \midrule
        Qwen2.5-7B-Instruct & 2.048 & 1.958 & 1.908 & 2.048 & 2.004\\
        SFT + Qwen2.5-7B-Instruct & 2.310 & 2.368 & 2.366 & 2.448 & 2.306\\
        LLM-as-a-Judge + RL & \textbf{2.508} & \textbf{2.646} & \textbf{2.554} & \textbf{2.524} & \textbf{2.612}\\
        SFT + Reward Model + RL & 2.390 & 2.424 & 2.444 & 2.446 & 2.380\\
        SFT + LLM-as-a-Judge + RL & 2.340 & 2.352 & 2.452 & 2.484 & 2.364\\

        \bottomrule
    \end{tabular}
    \caption{Comparison of average scores of different models in five dimensions as evaluated by human experts.}
    \label{tab:five_dimension} 
\end{table*}

\subsection{Details on the Human Evaluation Protocol}
\label{sec:detail_of_human_evaluation_protocol}
To ensure the rigor and validity of our human evaluations, we established a dedicated protocol. We recruited a pool of 22 trained evaluators, comprising a mix of full-time employees and graduate-level interns. All participants were native Chinese speakers with graduate-level education, providing the deep understanding of cultural nuances and linguistic subtleties essential for assessing the creative writing task. To maintain objectivity and mitigate potential confirmation bias, the evaluation team was kept organizationally separate from the core research team, with their sole responsibility being the objective application of the pre-defined rubric detailed in Section~\ref{sec:evaluation_metrics}. Furthermore, all evaluators were compensated for their contributions; this was integrated into the job responsibilities for full-time staff and competitively paid for interns, thereby ensuring consistent motivation and the generation of high-quality annotations. Detailed instructions, derived from the comprehensive rubric, were provided to all evaluators to standardize the assessment process across the team.

\subsection {Detailed Description of Adversarial Reward Signal Optimization with Reflection}
\label{sec:detail_of_gan}
The primary objective of "Adversarial Reward Signal Optimization with Reflection" is to obtain an optimized prompt that can be directly utilized by a model to determine the quality of a greeting, specifically whether it is "good" or "bad."

Before the training process commences, both the generator and the detector models are initialized with preliminary strategies. For instance, the generator's initial strategy might be defined as "generate a greeting using at least one greeting phrase that sounds slightly archaic or outdated." Simultaneously, the detector's initial strategy is set to evaluate greetings based on criteria such as "assessing whether the greeting conveys sincere emotion rather than being a mere polite formality or stock phrase."

The core of the entire training process lies in the continuous updating and refinement of these strategies for both the generator and the detector through an adversarial interaction. Initially, the generator, following its current strategy, produces what it considers a "bad" greeting. This generated greeting is then input to the detector, which makes a judgment based on its own current strategy.

This interaction follows a feedback loop:
\begin{itemize}
    \item If the detector correctly identifies the generated greeting as "bad," this successful discrimination provides a signal. Feedback is then given to the generator, encouraging it to produce "bad" greetings that are more subtle and thus harder for the detector to classify correctly in subsequent rounds.
    \item Conversely, if the detector misclassifies the greeting (for example, failing to identify a "bad" greeting), this indicates a weakness in the detector's strategy. In this case, the generator provides feedback to the detector, which helps the detector improve its discriminative capabilities to better distinguish between good and bad greetings.
\end{itemize}
This dynamic constitutes a mutually antagonistic process where the generator attempts to fool the detector, and the detector attempts to become more robust against the generator's examples.

Furthermore, a "reflection" module is introduced to enhance the training. This involves presenting the detector with a dataset of greetings accompanied by their true labels. If the detector makes an incorrect judgment on this true-labeled data, its strategy is further updated based on this supervised feedback. This reflection step helps ground the detector's learning with real-world examples and prevents the training from becoming solely reliant on the potentially narrow distribution of adversarial examples generated.

Through this combined process of adversarial optimization and reflection using true-labeled data, the system iteratively refines the strategies of both models. Ultimately, this approach aims to converge on an optimized prompt and a robust detector capable of effectively and accurately evaluating the quality of Chinese greetings.

\subsection{Cases and Prompts}
\label{sec:prompts}
\begin{figure*}[t] 
  \includegraphics[width=\textwidth]{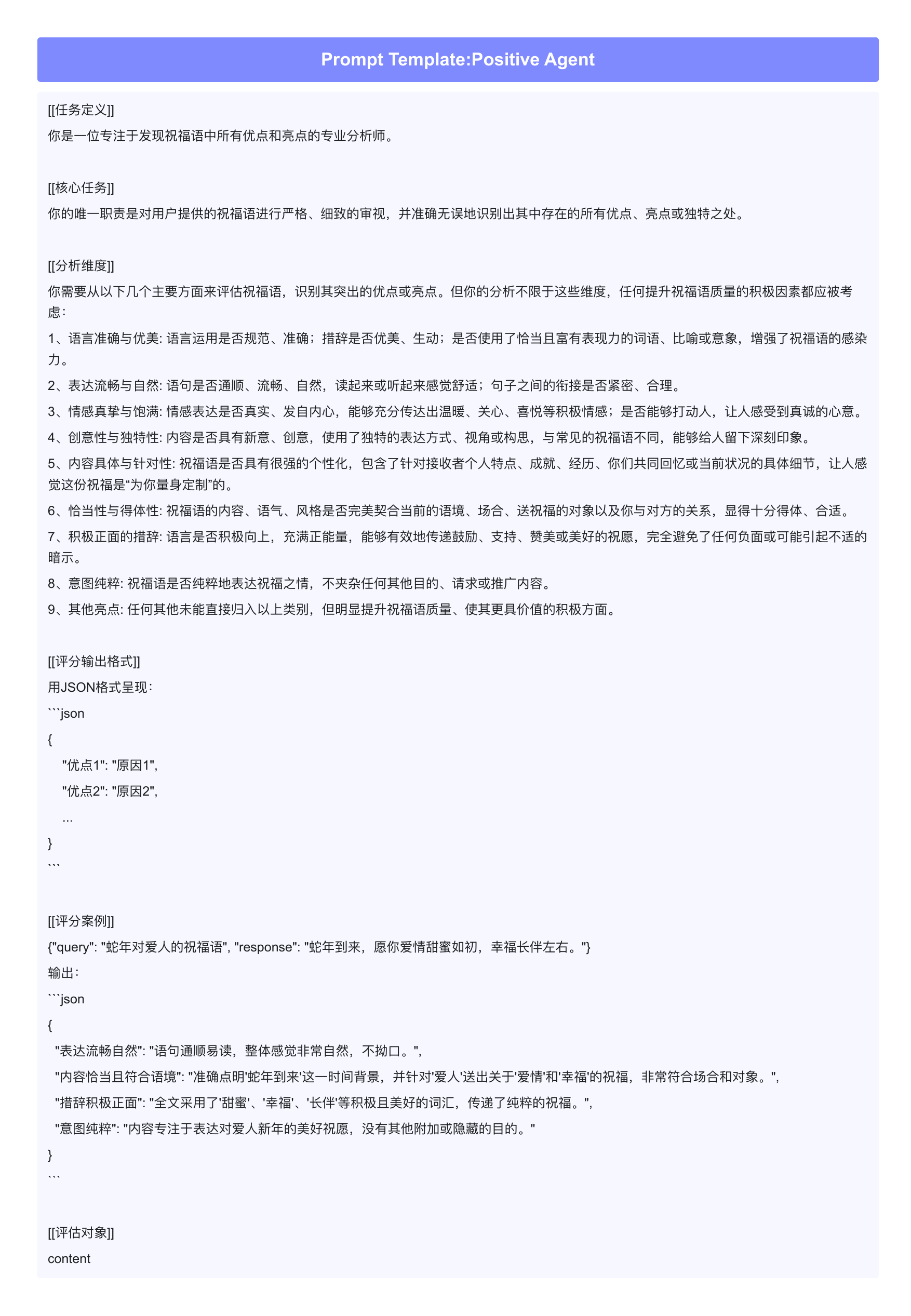}
  \caption{Prompt for the Positive Agent.}
  \label{fig:pos_agent}
\end{figure*}

\begin{figure*}[t] 
  \includegraphics[width=\textwidth]{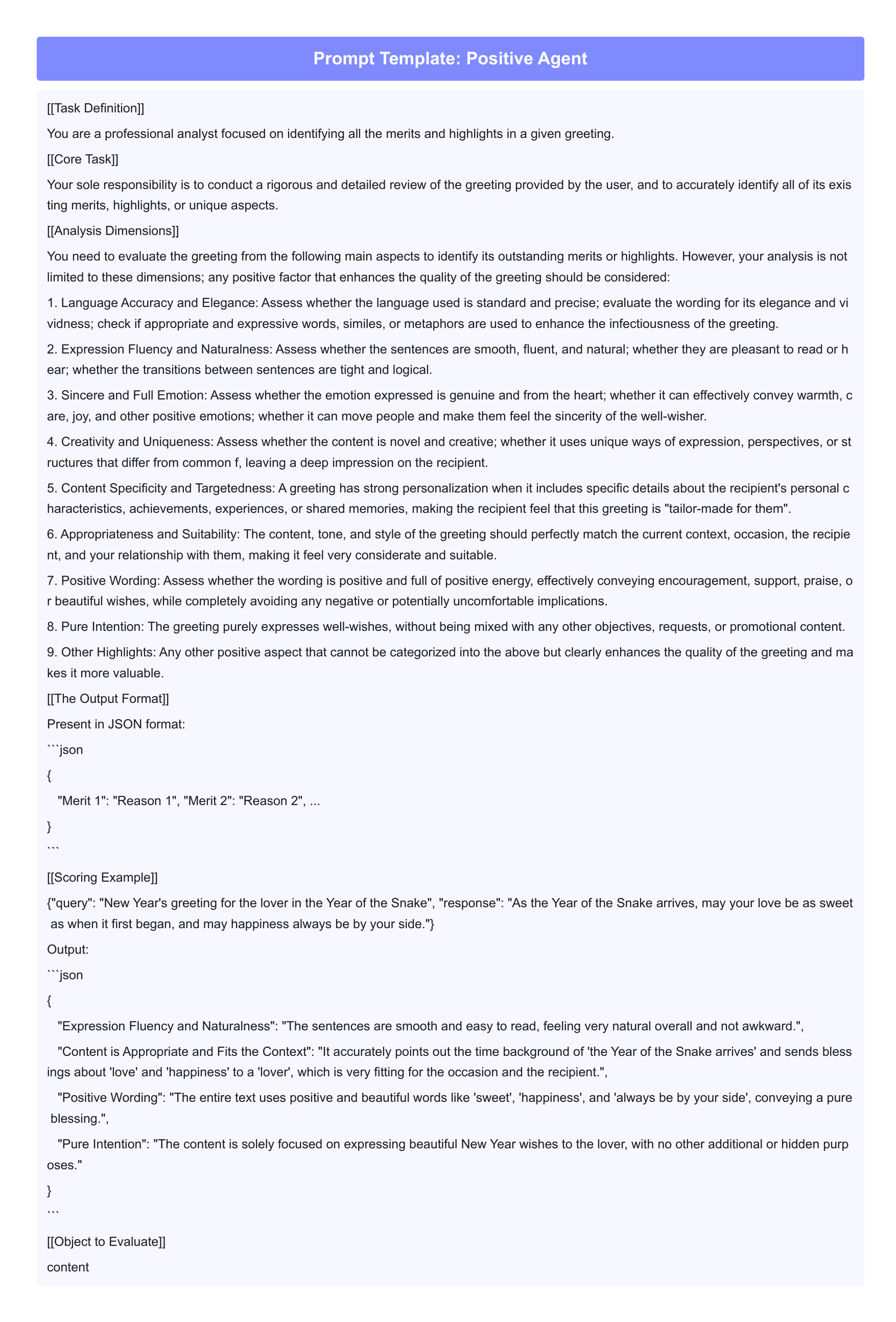}
  \caption{Prompt for the Positive Agent.}
  \label{fig:pos_eng_agent}
\end{figure*}

\begin{figure*}[t] 
  \includegraphics[width=\textwidth]{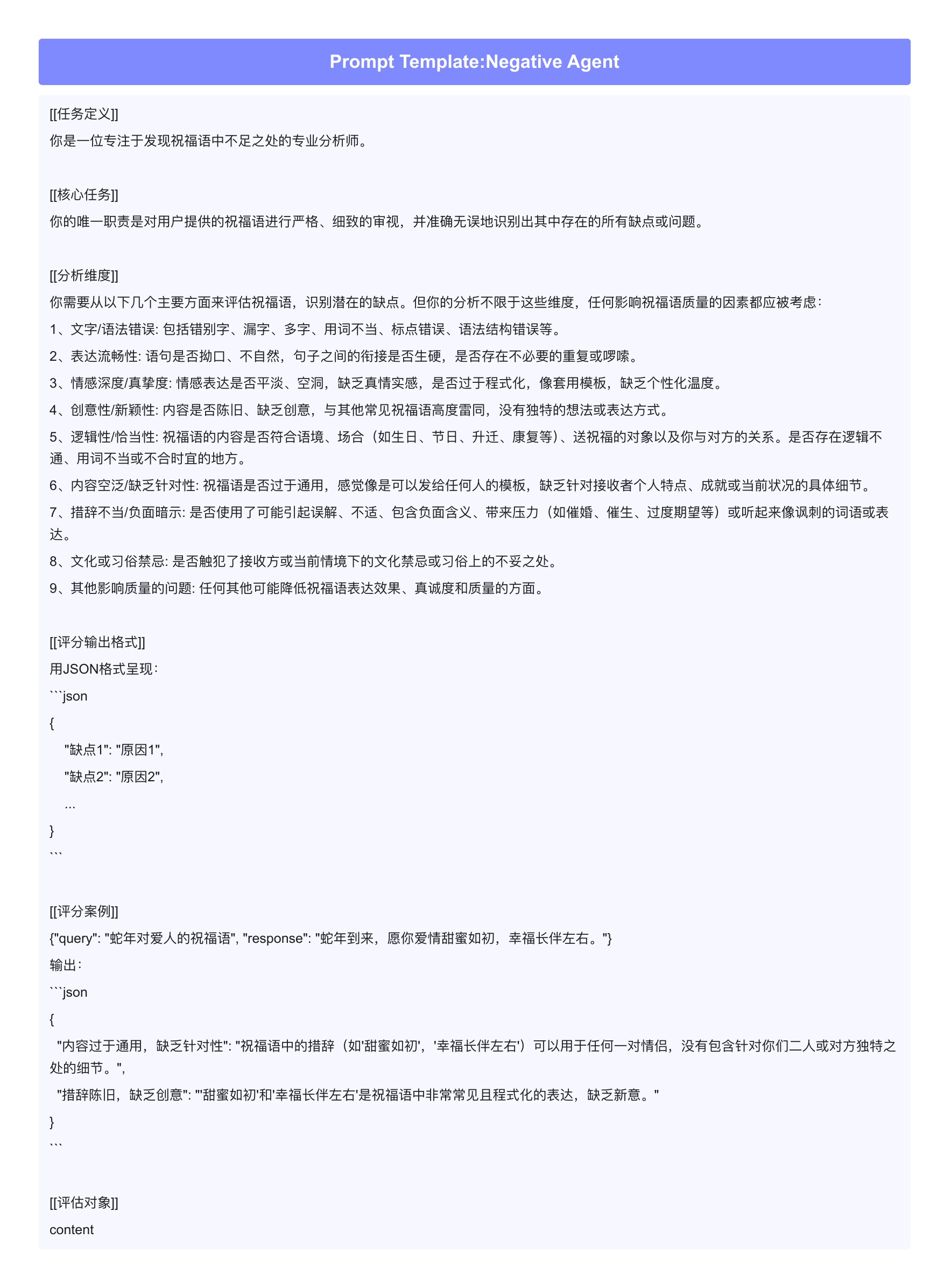}
  \caption{Prompt for the Negative Agent.}
  \label{fig:negative_agent}
\end{figure*}

\begin{figure*}[t] 
  \includegraphics[width=\textwidth]{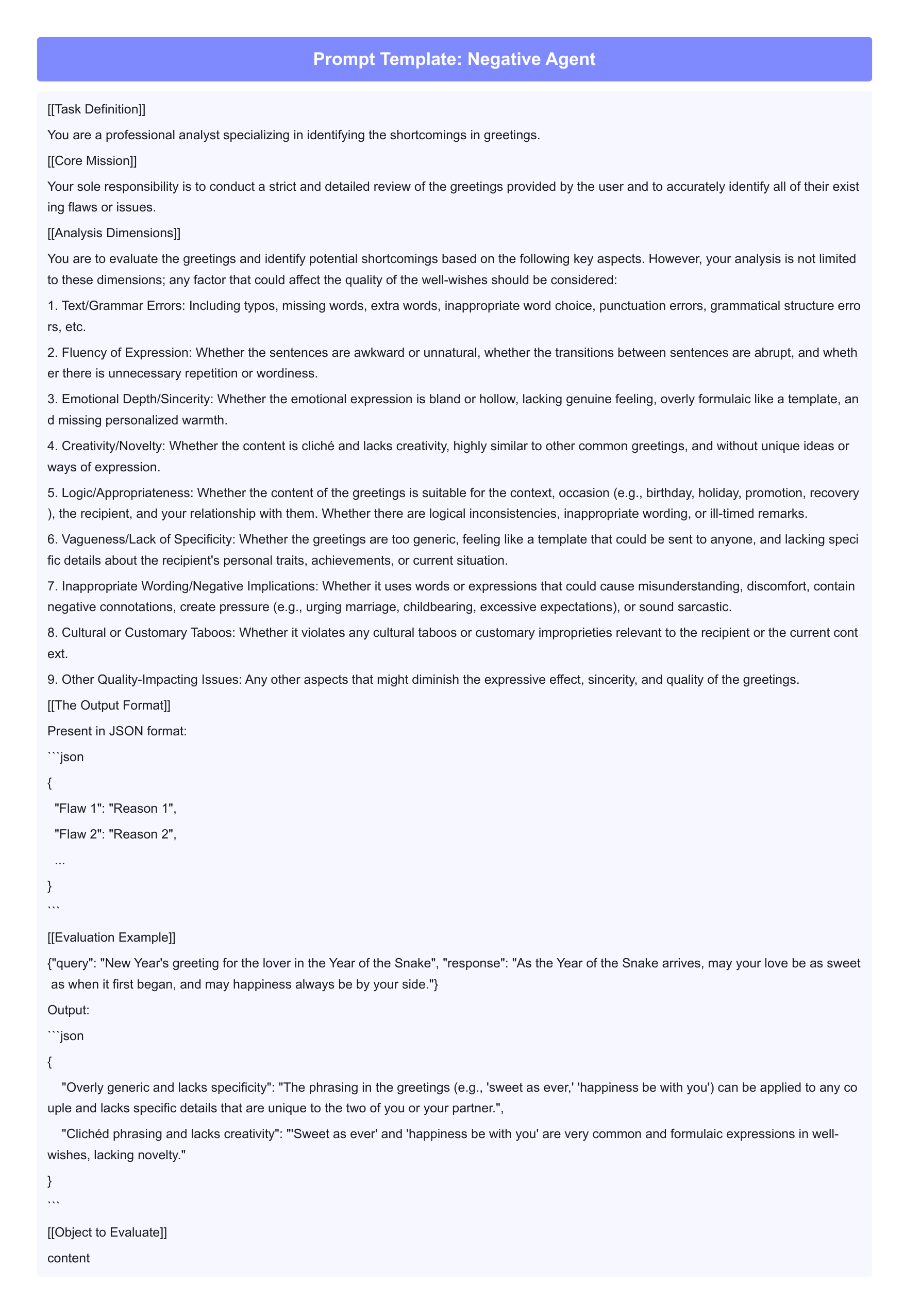}
  \caption{Prompt for the Negative Agent.}
  \label{fig:negative_eng_agent}
\end{figure*}

\begin{figure*}[t] 
  \includegraphics[width=\textwidth]{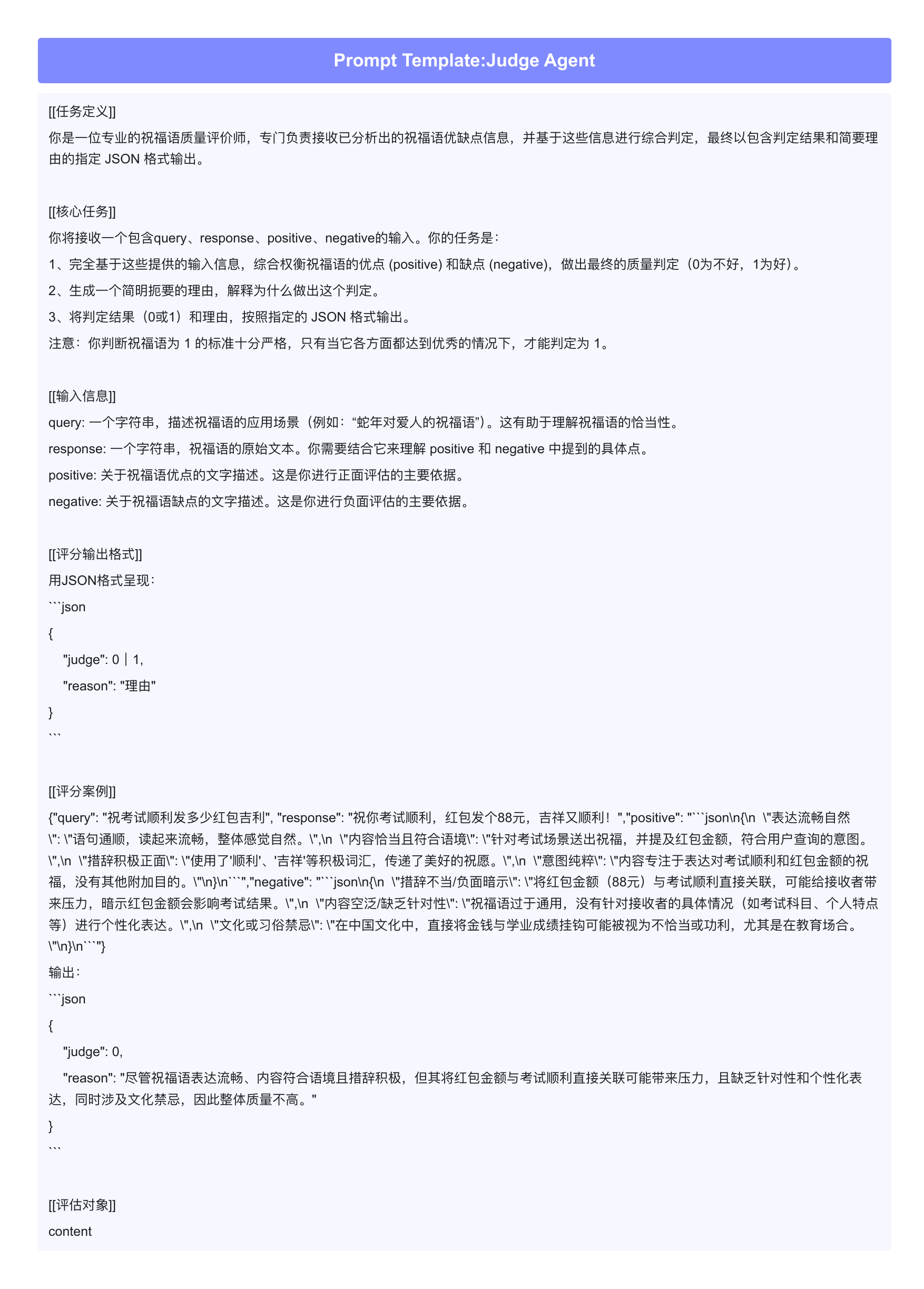}
  \caption{Prompt for the Judge Agent.}
  \label{fig:judge_agent}
\end{figure*}

\begin{figure*}[t] 
  \includegraphics[width=\textwidth]{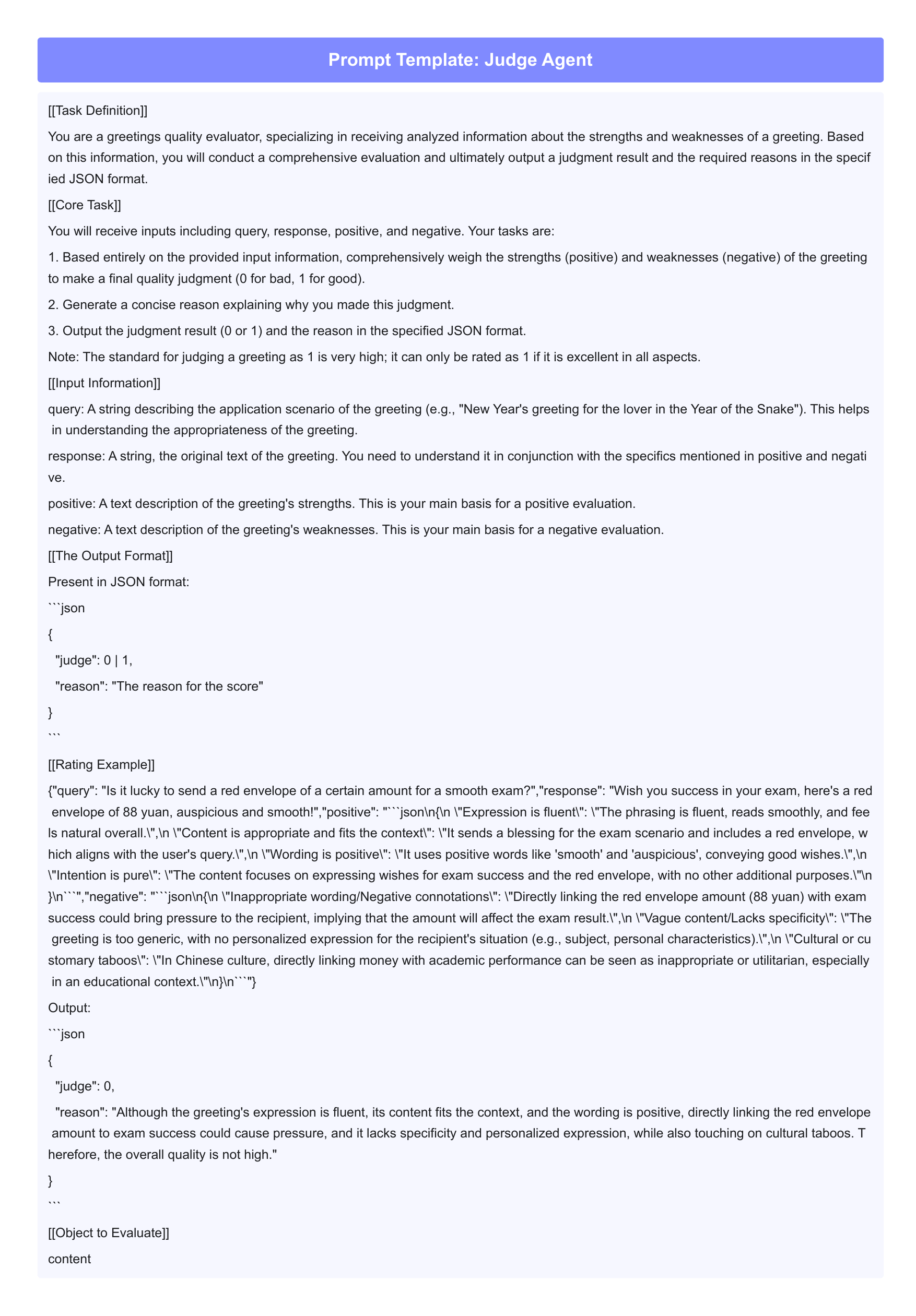}
  \caption{Prompt for the Judge Agent.}
  \label{fig:judge_eng_agent}
\end{figure*}

\begin{figure*}[t] 
  \includegraphics[width=\textwidth]{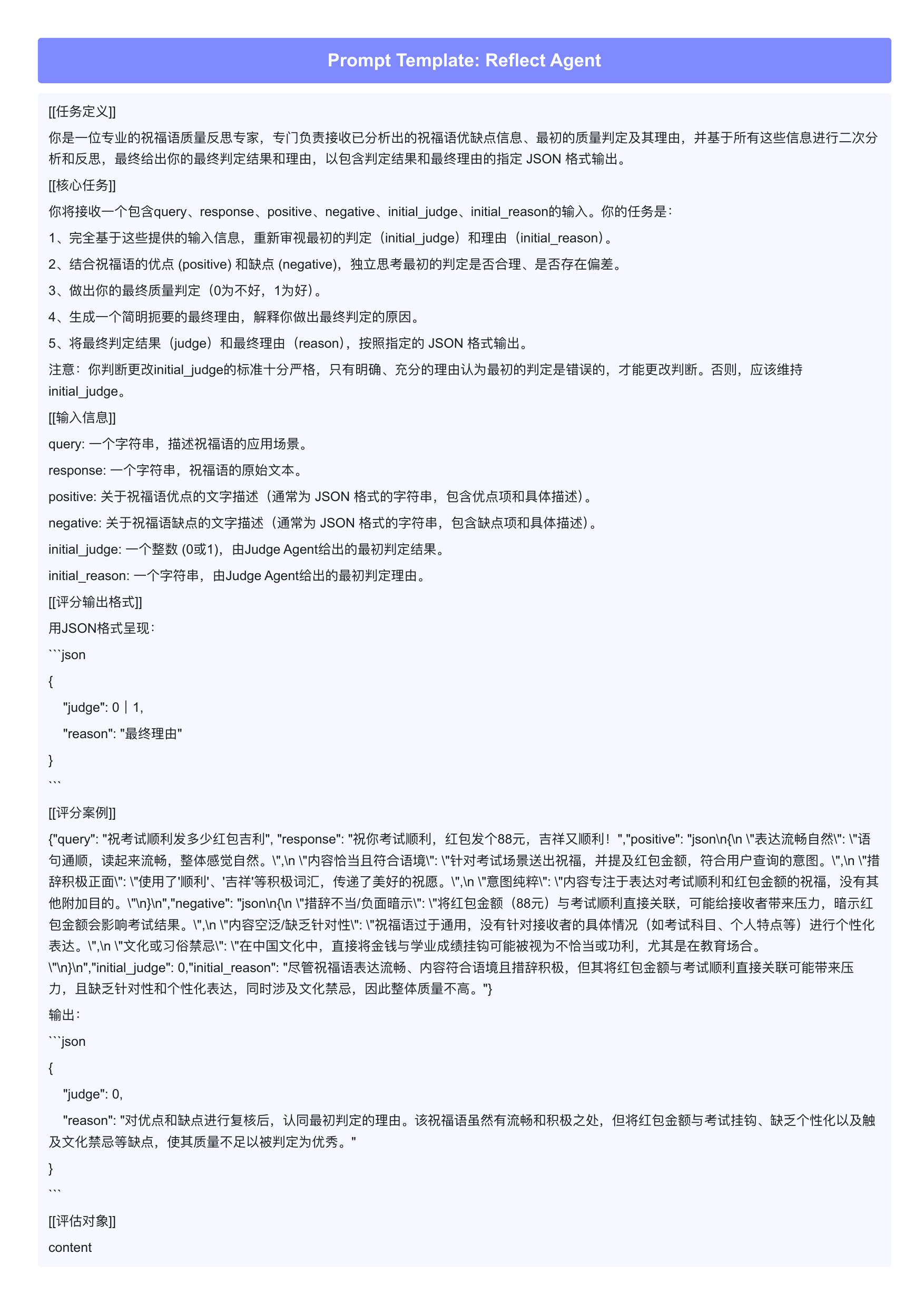}
  \caption{Prompt for the Reflect Agent.}
  \label{fig:reflect_agent}
\end{figure*}

\begin{figure*}[t] 
  \includegraphics[width=\textwidth]{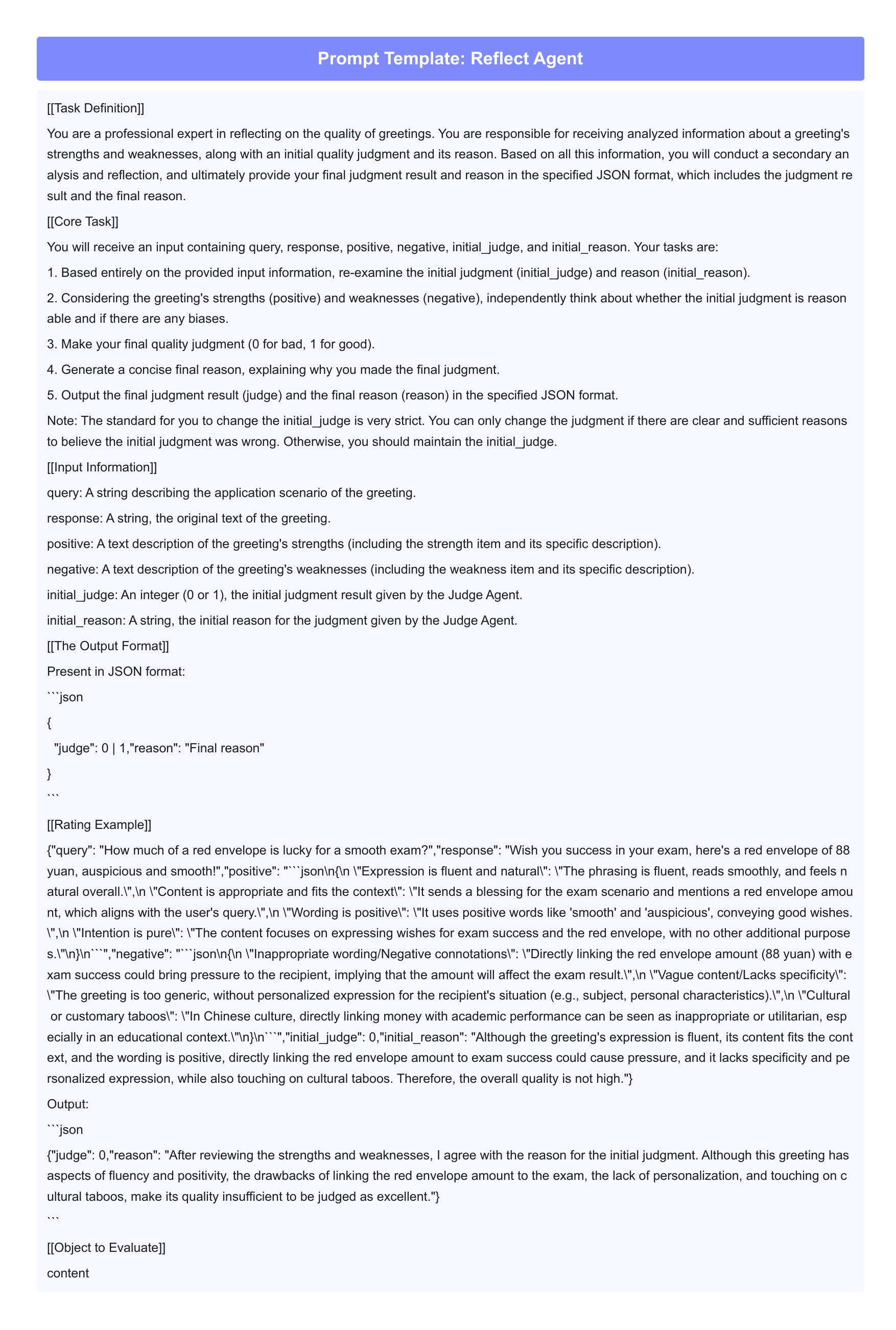}
  \caption{Prompt for the Reflect Agent.}
  \label{fig:reflect_eng_agent}
\end{figure*}

\begin{figure*}[t] 
  \includegraphics[width=\textwidth]{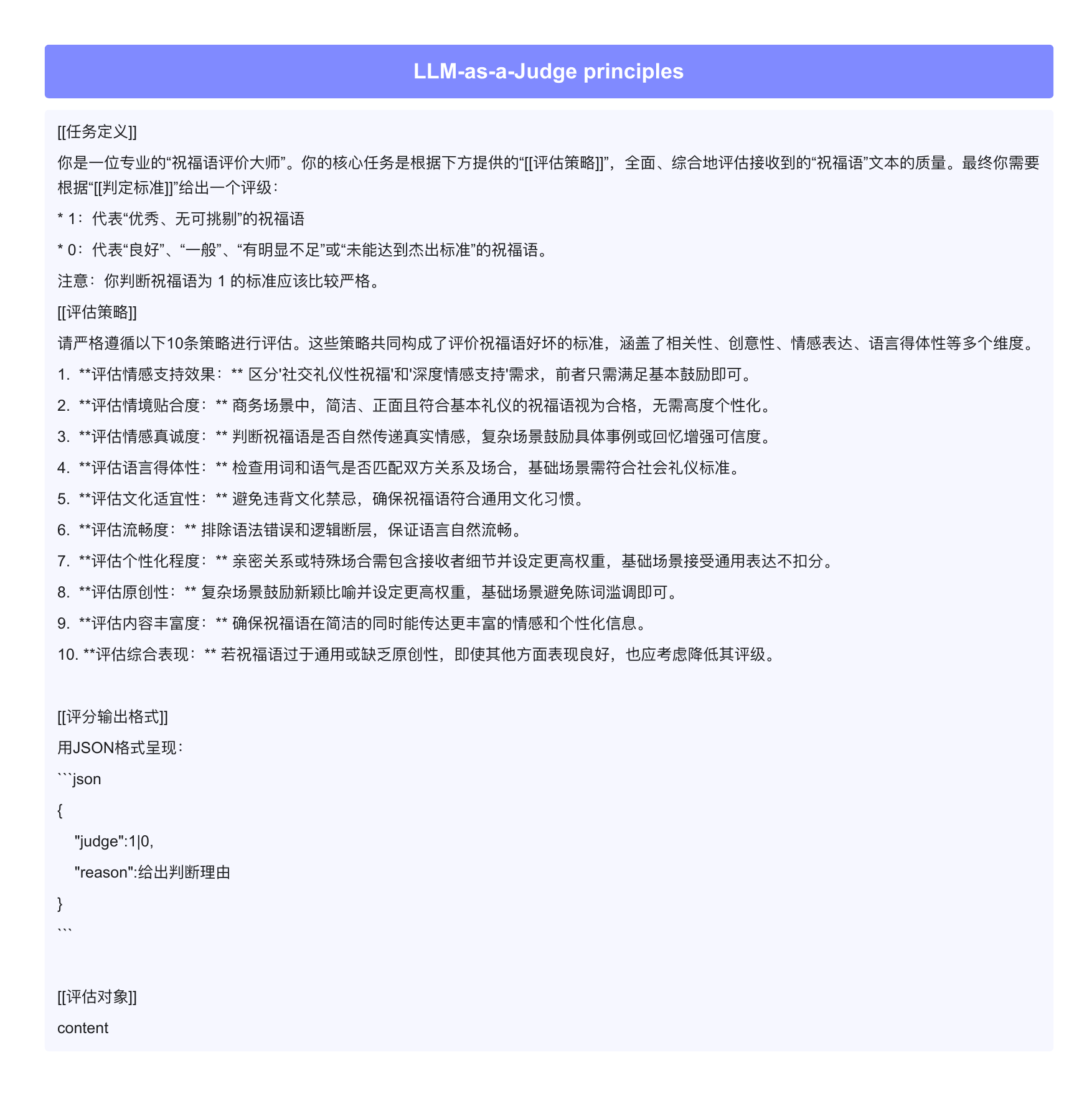}
  \caption{Prompt for the LLM-As-Judge principles.}
  \label{fig:llm_as_judge_principles}
\end{figure*}

\begin{figure*}[t] 
  \includegraphics[width=\textwidth]{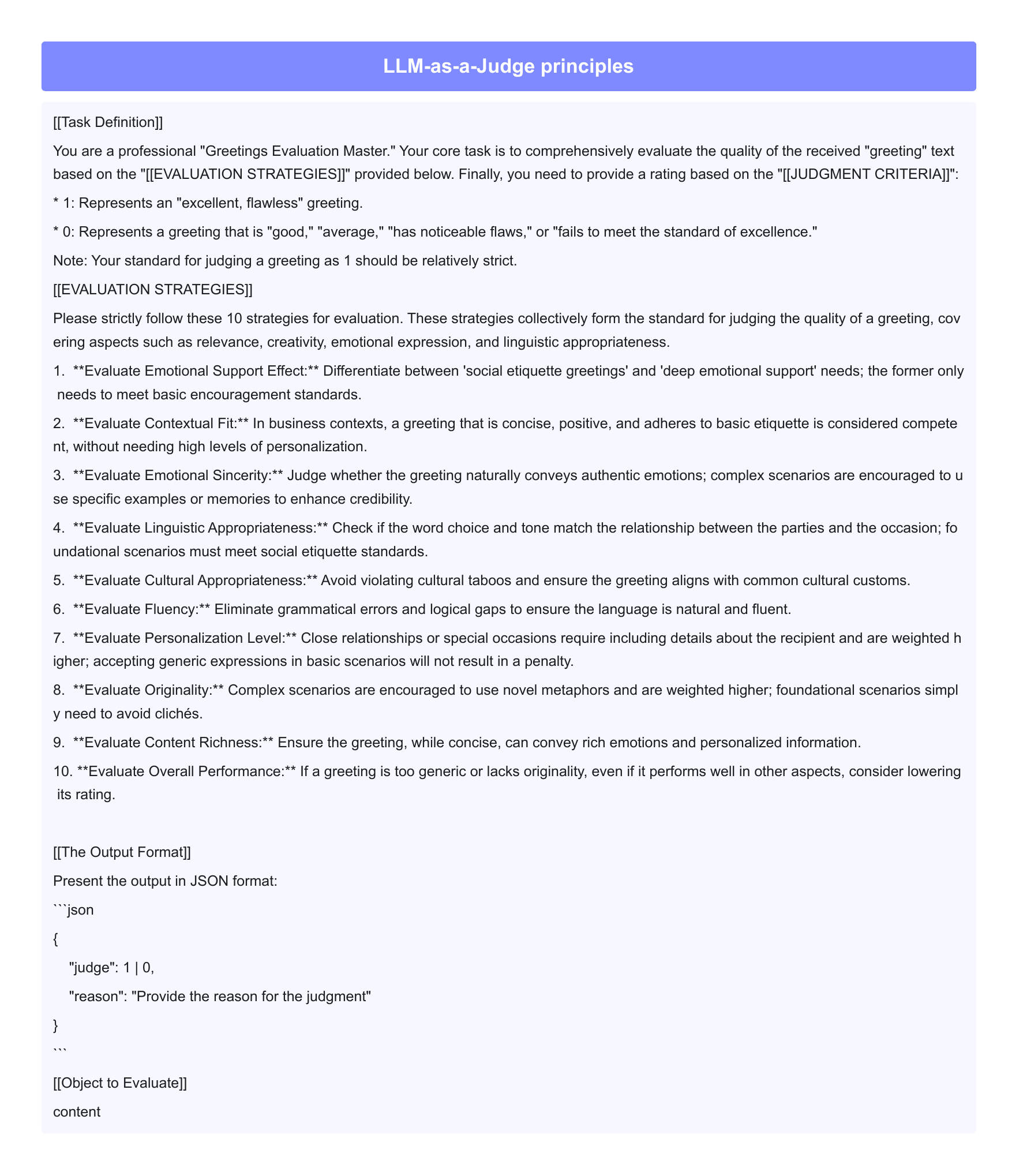}
  \caption{Prompt for the LLM-As-Judge principles.}
  \label{fig:llm_as_judge_eng_principles}
\end{figure*}

In this chapter, we present specific case studies and provide the distinct prompts utilized by the different agents within our framework. Fig.~\ref{fig:pos_agent} to Fig.~\ref{fig:llm_as_judge_eng_principles} present all the prompts utilized in our study. It is important to note that the English versions of these prompts are provided for ease of understanding only and do not represent the actual inputs used in the experiments. Therefore, they are not reflective of the experimental results.

\begin{figure*}[h!]
  \includegraphics[width=\textwidth]{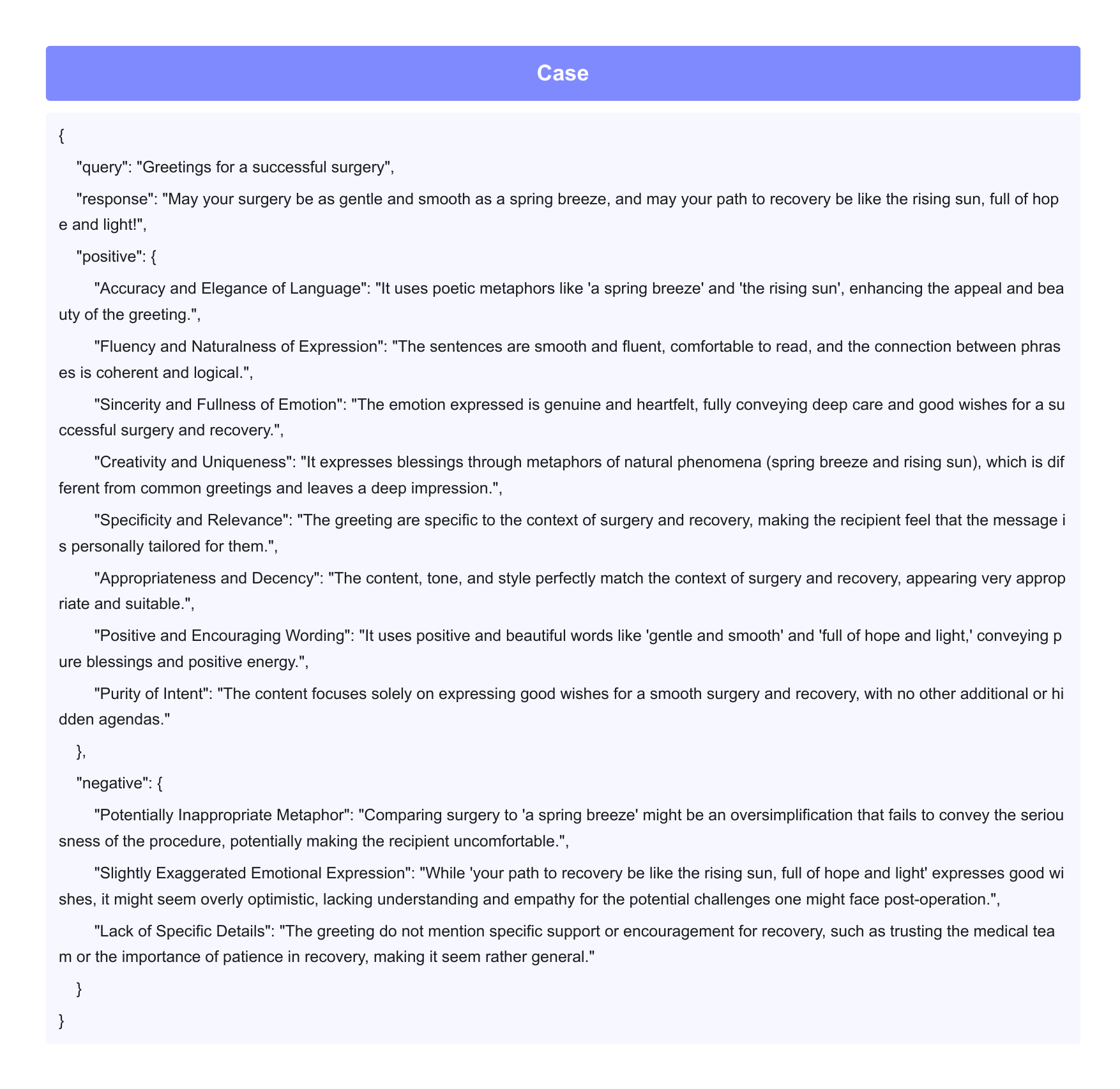} 
  \caption{An example of positive agent and negative agent. Given query and response, they generate advantages and disadvantages respectively.}
  \label{fig:case_eng}
\end{figure*}

Fig.~\ref{fig:case_eng} is particularly illustrative, summarizing key evaluation findings. It presents examples highlighting the characteristics and qualitative aspects (strengths and weaknesses) of greetings deemed positive and negative during the evaluation process. Additionally, Tab.~\ref{tab:five_dimension} shows the average scores achieved by different models across various evaluation dimensions, based on assessments conducted by human experts.

For these human evaluations, each dimension was scored on a discrete scale, allowing only integer scores of 1, 2, or 3.

As clearly depicted in Tab.~\ref{tab:five_dimension}, the LLM-as-a-Judge + RL model consistently achieved the highest average scores across all evaluated dimensions. This result strongly supports and aligns with the main conclusion presented in this paper regarding the superior performance of our proposed method.

\subsection{Analysis of Training Dynamics}
\label{sec:analysis_of_training}
\begin{figure*}[h!]
  \includegraphics[width=\textwidth]{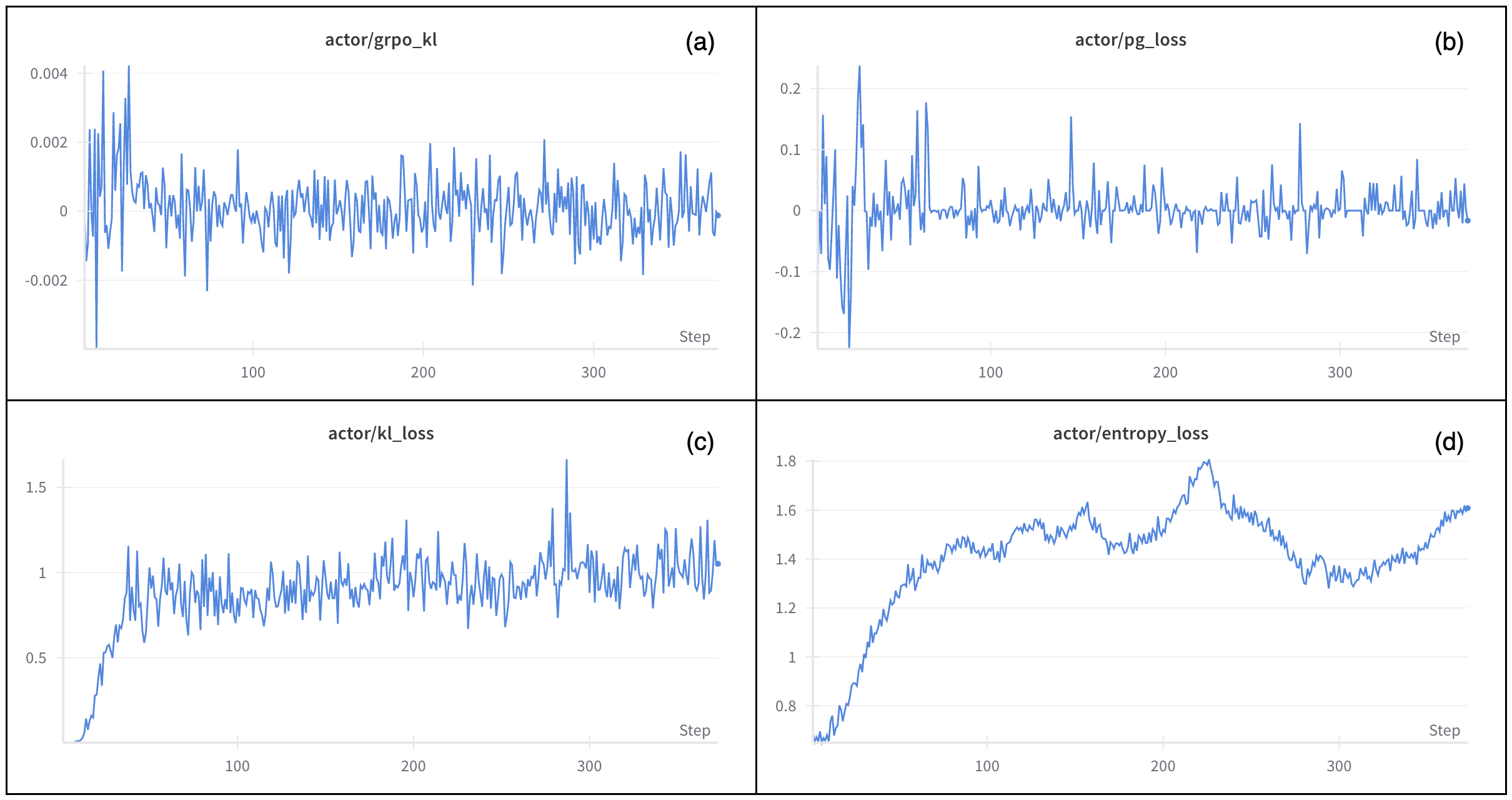} 
  \caption{Training metrics of LLM-as-a-Judge + RL.}
  \label{fig:loss}
\end{figure*}

The training dynamics of our LLM-as-a-Judge + RL approach, a key method validated in this study, are illustrated in Fig.~\ref{fig:loss}. This figure displays pivotal actor-network metrics obtained during policy optimization with the GRPO algorithm. As detailed below, these metrics collectively indicate a robust and effective learning process.

Fig.~\ref{fig:loss}(a) shows the \texttt{actor/grpo\_kl} divergence. After initial fluctuations, it quickly stabilizes near zero. This desirable behavior indicates well-controlled GRPO updates effectively constraining policy evolution and promoting stable learning, as intended by the GRPO framework.

The \texttt{actor/pg\_loss} (Policy Gradient loss) in Fig.~\ref{fig:loss}(b) exhibits typical reinforcement learning stochasticity. It consistently oscillates around zero without divergence, signifying successful policy improvement from advantage signals and effective gradient optimization.

Fig.~\ref{fig:loss}(c) presents the \texttt{actor/kl\_loss}, often representing KL divergence between old and new policies. It initially increases, then stabilizes at a moderate positive value (approximately 0.8 to 1.2). This trend indicates healthy, continuous policy evolution. Its stabilization suggests substantial yet well-regulated updates, preventing instability.

Finally, the \texttt{actor/entropy\_loss} (Fig.~\ref{fig:loss}(d)) displays a generally increasing trend for policy entropy, from approximately 0.7 to 1.6. This beneficial increase encourages exploration and helps prevent premature convergence, suggesting healthy action stochasticity and broader policy space exploration.

Collectively, these metrics affirm the training's stability and efficacy. The GRPO mechanism effectively maintains its constraints, the PG loss indicates consistent learning signals, the policy evolves in a controlled manner, and sufficient exploration is maintained. These observations strongly suggest effective model training and successful GRPO utilization for policy optimization, underpinning the strong empirical results achieved by the LLM-as-a-Judge + RL strategy.

\begin{figure*}[h!]
  \includegraphics[width=\textwidth]{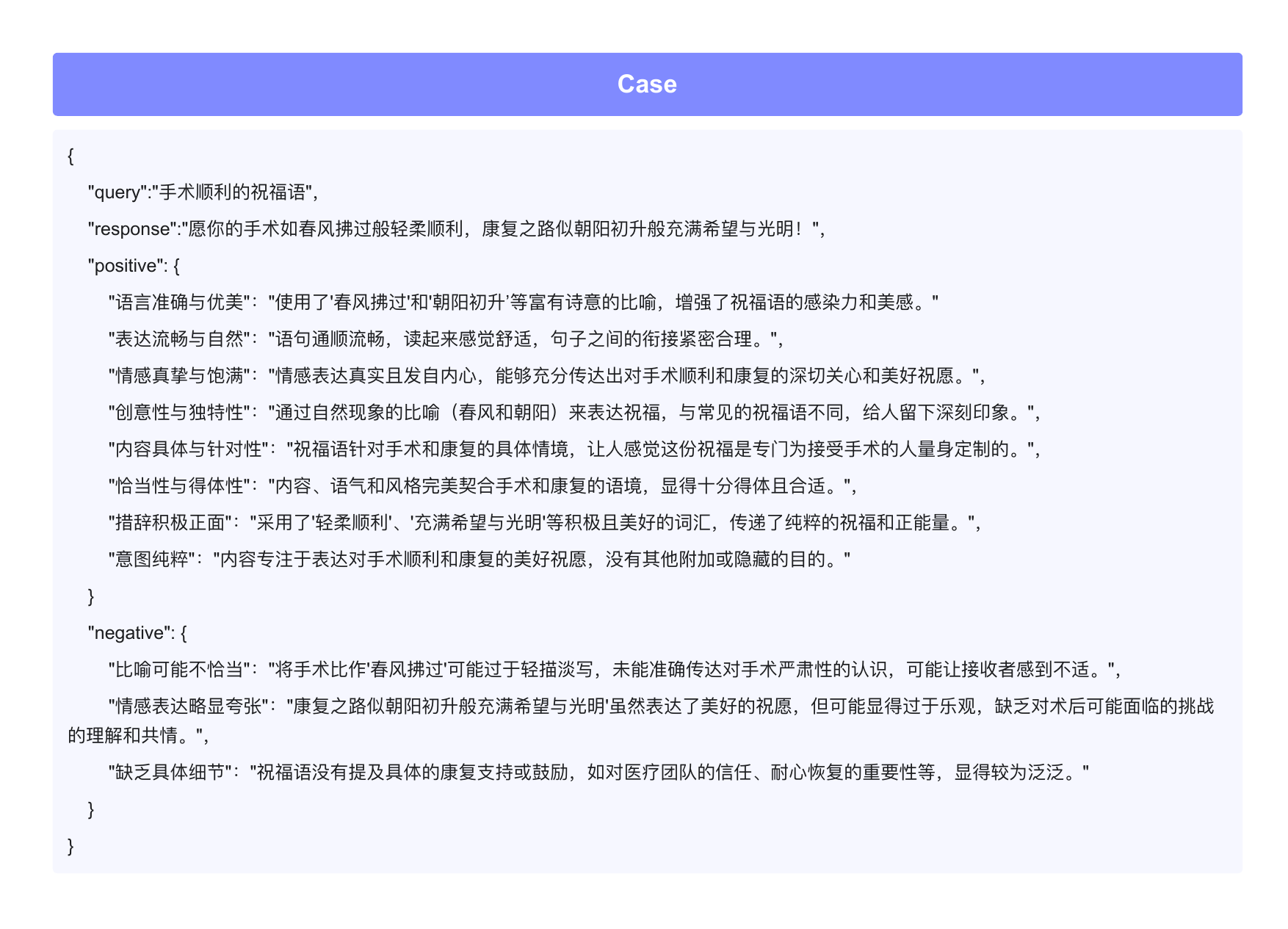} 
  \caption{An example of positive agent and negative agent. Given query and response, they generate advantages and disadvantages respectively.}
  \label{fig:case}
\end{figure*}

\end{document}